# Increasing Resilience of Continuum Robots via Motion Planning Algorithms

Oxana Shamilyan[1*[0000-0002-0961-9059]], Ievgen Kabin[1], Zoya Dyka[1,2], Oleksandr Sudakov[3] and Peter Langendoerfer[1,2]

[1*] IHP – Leibniz-Institut für innovative Mikroelektronik, Frankfurt (Oder), Germany
[2] BTU Cottbus-Senftenberg, Cottbus, Germany
[3] Technical Center, National Academy of Sciences of Ukraine, Kyiv, Ukraine

**Abstract.** This paper presents an experimental study of motion planning for resilient continuum robots. In this study we mainly focused on multi-criteria decision-making, its application for path-planning algorithms, impact on the generated path and execution time. To do this, we used two well-known algorithms for path planning, namely Genetic algorithm and A* algorithm, and modified them by adding the Analytical Hierarchy Process algorithm to evaluate the quality of the paths' generated. In our experiment the Analytical Hierarchy Process considers four different criteria, i.e. distance, motors damage, mechanical damage of the robot's arm and accuracy, each considered to contribute to the resilience of a continuum robot. The use of different criteria is necessary to increase the time to maintenance operations of the continuum robot. We conducted the experiments using two different simulated environments of the robot. Although we significantly simplified the robot's model and its environment, we still implemented some of the features of the environment based on the real robot prototype. In particular, one of the environments has single- as well as multipath points, and other consists of the multi-path points only. The results show that, in contrast to A*, the performance time of Genetic algorithm does not depend on the environment's cardinality. It generates more diverse paths, which increases the robot's resilience.

## 1 Introduction

Continuum robots are a new type of robots providing a high degree of flexibility when it comes to reaching tricky positions, as they are not limited in their motions by joints. Instead, they have infinite degrees of freedom. On the one hand, this makes them ideal candidates for application areas such as inspection tasks, handling of unstructured objects, and minimally invasive surgery [1, 2]. On the other hand, the high flexibility of these robots presents a challenge in terms of controlling their motions. Continuum robots are often used in environments in which flexibility is required. For such tasks the continuum robots need to navigate, move and make decisions about next moves. Environments in which continuum robots are used may be difficult to access and/or dynamically changing and/or require precise motions. These complex environments require autonomously operating robots. Autonomous operation of continuum robots in complex environments is still a challenge. Thus, such robots are often operated by humans, and problems of resilient communication between humans and robots arise. There is a lot of research focusing on the automatic control of the robots, providing a comprehensive survey of the most popular continuum robot types, motion control models, and examples of research groups that are particularly focused on the study of motion control problems of the continuum robots [2].

The ability to generate and choose the path in the unknown environment is one of the first steps towards achieving robot autonomy. This feature is necessary for the implementation of tasks for continuum robots, which are related to the detection and elimination of problems that lead to inoperability of the inspected system (alive or mechanical). Autonomy is one of the features of the modern resilient systems.

The core features of resilience are security, reliability and autonomy. Different cryptographic protocols and algorithms can be used for implementing data protection and for realizing security goals. Redundancy is the usual means to reach reliability. Autonomy, in a very narrow sense, is often defined as independence from a power source. In our research we are focussing on autonomy but our definition autonomy includes more aspects than independency to energy supply. In our definition, autonomous systems are required to be at least:

- self-aware,
- robust to fluctuation of environmental parameters (or self-adaptable),
- resistant against malicious manipulation,
- intelligent enough to find correct solution(s) in complex and unexpected situations on their own [3].

The implementation of these features is not a simple or straightforward process. In this paper, we focus only on path planning for continuum robots. The path planning task can be a part of self-adaptability as well as a reaction to unexpected changes in the environment or internal state parameters. Under these conditions the path planning should take into account more parameters than only the start and the goal coordinates. Using environmental and internal state parameters as additional criteria for path planning helps the system to make decisions and generate the optimal path. Here we mean by optimal not essentially the shortest path but the one with the least negative impact on the continuum robot. In order to enable a robot to properly assess more conditions than just the path length, we have modified path planning algorithms and added an Analytical Hierarchy Process (AHP) decision-making algorithm. The latter generates weights to the defined criteria and evaluates the generated path, to find the optimal solution. In our experiments we used Genetic algorithm (GA) and A* algorithm for path planning.

This paper is an extended version of [4] that was previously presented by the authors at the Multi-Objective Decision Making Workshop (MODeM) at the 27th European Conference on Artificial Intelligence (ECAI 2024).

We extended our previous paper with the following aspects:

1. expanded definitions of the path planning algorithms, decision-making algorithm as well as the defined criteria,
2. experiments with the different configurations of the continuum robot,
3. analysis of each single criterion of the decision-making algorithm,
4. new results, compared to the short version of the paper, regarding the performance of GA and A* algorithm.

The rest of this paper is structured as follows. Section 2 introduces our physical robot prototype, its abilities and technical parameters. Section 3 describes the simplified robot model as well as two different simulated environments. Section 4 provides information about the algorithms explored in this study: the multi-criteria decision-making algorithm AHP and the two path planning algorithms GA and A*. Section 5 discusses the conducted experiments and obtained results. This paper ends with a short conclusion.

## 2 Physical prototype and its motions

The experiments with path planning and decision-making algorithms were planned to be run on a self-made tendon-driven continuum robot prototype. This section provides a description of the robot prototype, its abilities and technical details.

### 2.1 Prototype of the Continuum Robot

Our self-made tendon-driven continuum robot prototype is shown in Fig. 1. It consists of 15 disks (30 mm diameter and 1 mm height), one central flexible backbone tendon (3 mm diameter), and 8 side tendons (1 mm diameter). The disks are rigidly fixed on the central backbone tendon at a distance of 20 mm from each other and they do not change their position relative to the central backbone. The prototype has two flexible sections (lower and upper section, showed with white arrows in Fig. 1). Each section has a base (the bottom disk) and a tip (the top disk). We assume that the robot's tip is located in the centre of the top disc. The lower section has a fixed base and is considered as the robots' base. The base of the upper section is mobile and coincides with the tip of the lower section (see "connection point" in Fig. 1). The tip of the upper section is considered as the robot's tip. Side tendons pass through each disk along both sections (4 side tendons for each section), so that each section can be controlled independently using stepper motors [1], [2].

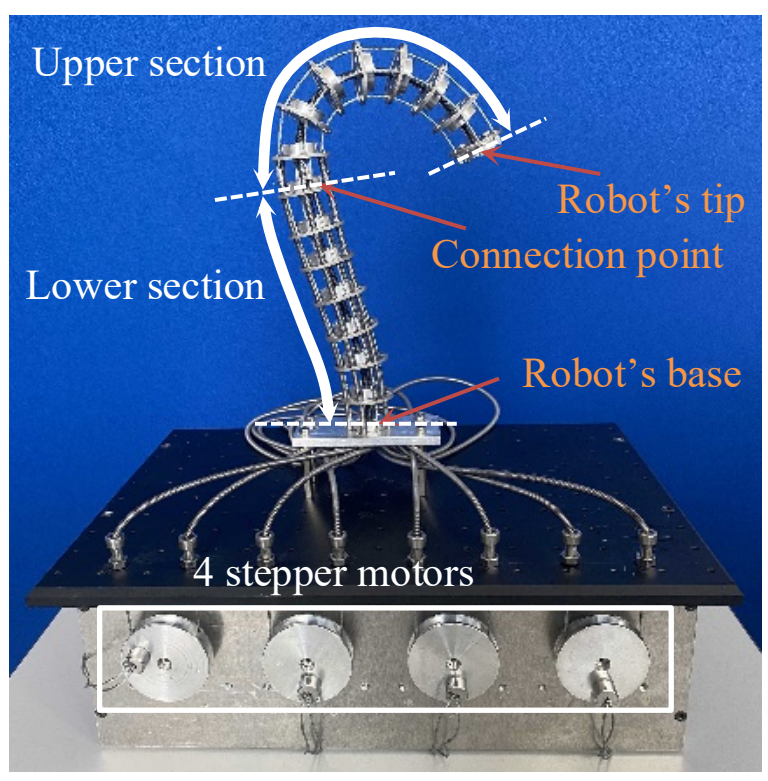

**Fig. 1.** Self-made tendon-driven continuum robot prototype

We used 2 stepper motors NEMA23-31-01SD for each section, i.e. the prototype is actuated by 4 motors. Please note that the use of stepper motors reduces the flexibility of the robot. Each motor needs 800 steps per revolution that makes the robot’s working environment no longer analogue but discrete[1], i.e. consisting of a finite number of points. In the rest of the paper we denote the set of all points that a single section of the robot can reach with its tip as “local environment” of the section. The set of all points that a robot using both sections can reach with its tip is denoted as “global environment” of the robot. If the local environment consist of n points the global environment of the 2-sections robot consists of $n^2$ points, if both robot’s sections are identical.

The stepper motors are connected to the microcontroller board ARD MEGA2560R3 with an ATmega2560 microcontroller that controls the rotations of the stepper motors. The connection of the motors to the microcontroller prevents their simultaneous use. As both motors of a section are connected to the same microcontroller, they can only be used in series, as the microcontroller does not provide parallel processing. The stepper motors are equipped with absolute encoders AMT222A-V, which count their steps. Motions of our robot prototype can be controlled either manually or by software. For the manual control two joysticks are used, one for each section of the robot. For the programmed (automated) control a path planning calculation program can be used to determine the path before each new motion. Path planning calculations are executed on a laptop equipped with an Intel Core i5-10310U CPU @ 1.70GHz Processor and 8 GB RAM. The microcontroller board and the laptop communicate via the UART protocol, while the encoders and board utilize the SPI protocol. The joysticks are connected to the microcontroller board via USB. A simultaneous use of encoders and joysticks is not possible, due to incompatibility of simultaneous use of USB and SPI ports in the microcontroller board.

[1] The number of steps per revolution means the number of steps the motor shaft does to complete one full turn (360°). 800 steps per revolution means that for each step the motor shaft turns at an angle of 0.45°.

### 2.2 Robot Motions Description

The description of the continuum robots' motions is a complex task. This type of robots has infinite degrees of freedom. Continuum robots are highly flexible, can easy deform and return to their original shape. That makes it hard to describe the robot's motions and calculate the trajectory of the robot and coordinates of the robot's tip. This problem has been described in many works [5–9]. The Cosserat rod model is a popular method to describe the motions of continuum robots. A Cosserat's rod is represented as a long, thin and perfect flexible body. The Cosserat's rod model considers each deformation of the rod caused by some external force ***F*** and allows to precisely calculate the resulting shape and position of the deformed body. In Fig.2 an example of the deformation is shown: initially, the rod was straight, not deformed and horizontal. The external force ***F*** is applied to the rod and deforms it. Vector ***d*** defines the normal of the cross-section. The Cosserat model allows calculating the deformed shape of the robot depending on the applied force, at the point of its application. More information about the Cosserat's rod model and its application examples can be found in [2].

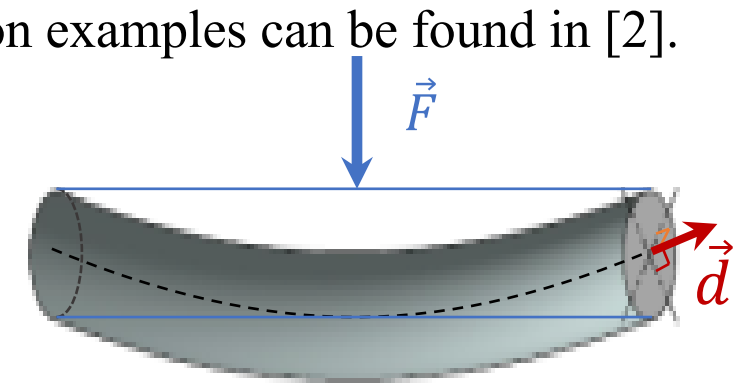

**Fig. 2.** The Cosserat rod model – the rod in a deformed state under the influence of force ***F***.

The motions of the robot are first of all depending on the robot's design. Instead of the force ***F*** which is shown in Fig.2 the deformation of our physical prototype is caused by changing tendons lengths due to rotation of the stepper motors. Fig.3 shows 1 section of our prototype schematically. The section has 2 tendons – "blue" and "orange". Each tendon is attached to one side of the robot's tip disc, runs down along the robot's section, is wound around the motor's shaft, and then runs up the robot's section and is attached to the robot's tip disc opposite to the other end, for example the blue tendon shown in Fig.3(a) is connected to motor 1 (later in this paper we use terms "motor 1" and "motor 2" to indicate the shaft of the motor 1 and 2, respectively). Another tendon, independent of the first one, is connected to the second motor and runs along the robot's section in an orthogonal plane to the first tendon, see orange tendon in Fig.3(a). The robot's motions can be described as bending that is caused by changing the lengths of the tendons parts, due to the rotation of the motors. When the motor rotates, it pulls on one end of the tendon and releases the other end, resulting in a bend. Thus, one motor bends the robot's section in the *x0z* plane (motor 1, "blue tendon"), and the second one – in the *y0z* plane (motor 2, "orange tendon"). Fig.3(a) shows the robot in its initial state, i.e. in a straight up position. The left and the right parts of the "blue" tendon have the same length denoted as l. The length of the both parts of the "orange" tendon is also l. Fig.3(b) shows a deformed robot after motor 1 caused a bending in the *x0z* plane. The robot's fixed base prevents the robot from rotating around the *z* axis. The number of motor steps that our robot can make along one axis from its most rightward to its most leftward bending position is equal to $n$=700 steps (350 steps to the left and right side from the initial state), i.e. a little less than one full revolution (i.e. 800 motor steps) of the motor shaft. The robot's tip can move in 4 different directions from every point in its environment, except of the edge points.

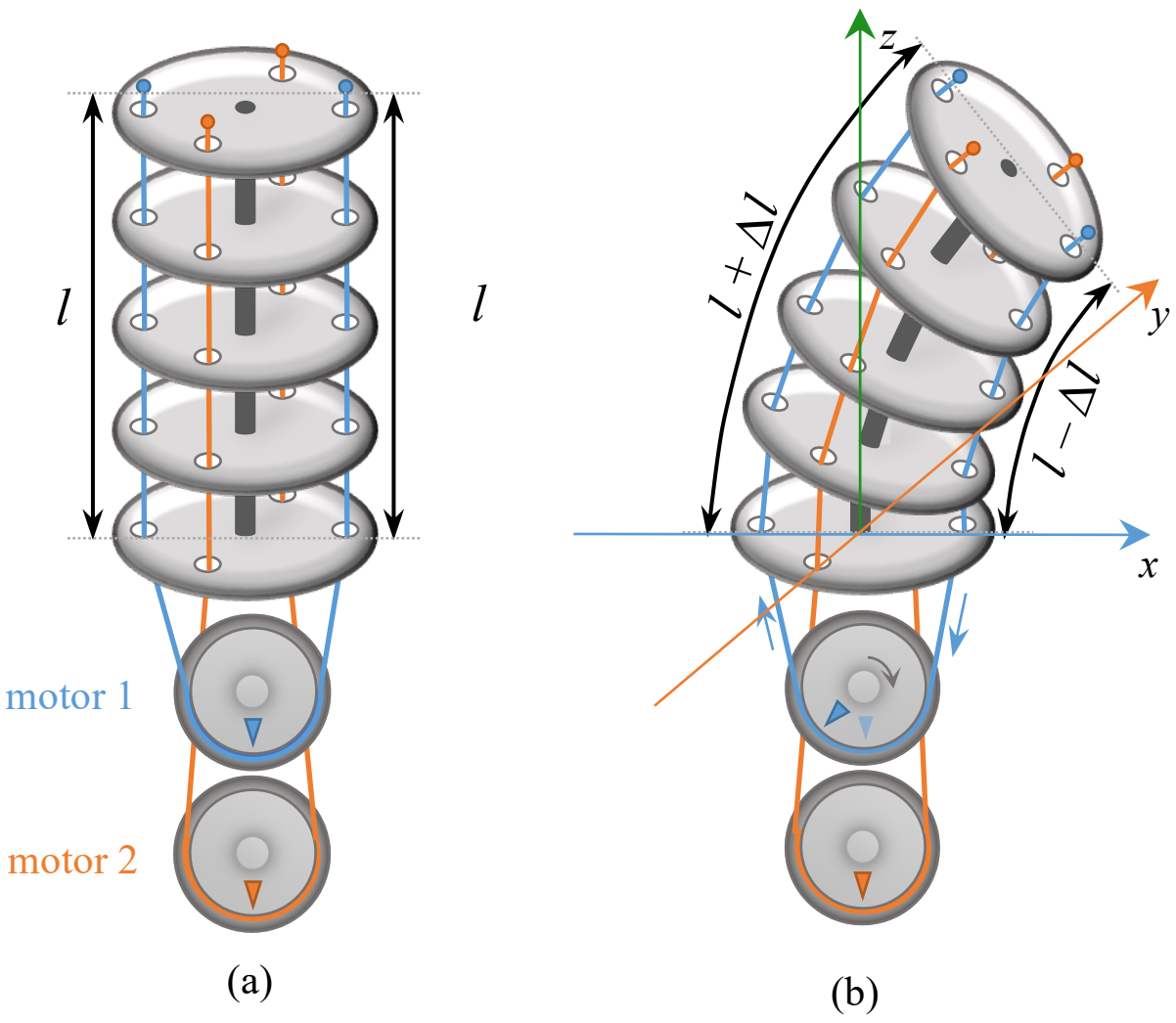

**Fig. 3.** Schematic representation of one-section of our robot and the position of the tendons: (a) robot in the initial state, (b) robot in a deformed state: the deformation is caused by the rotation of motor 1 that changes the lengths of the left and right parts of the "blue" tendon.

The length by which one side of the tendon decreases after motor rotation is denoted in Fig.3(b) as $\Delta l$. Simultaneously, the other side of the tendon increases by $\Delta l$. Knowing the length of the central backbone tendon l, which remains constant, as well as the new lengths of the side tendons $l+\Delta l$ and $l-\Delta l$, and the distances between the central and the side tendons, the trajectory of the motions and the coordinates of the robot tip can be calculated for the robot's section without implementing Cosserat's rod model, i.e. by focusing only on the type of deformations described.

## 3 Simplified robot model and its environments

In this work, we focus on the analyses of multi-criteria decision-making and its application for path-planning algorithms. For these purposes, we experimented not with the physical robot prototype but with a simulated robot model and its environments. To describe the motion of continuum robots, some researchers used Cosserat's rod model [5–9]. Alternatively, video records with markers to track the robot's motions were applied [10]. We do not use the exact shapes that the robot's prototype takes as it moves in our simulations. For our investigations, we simplified the robot model as well its environments. This section describes the main assumptions which we applied for the simplified robot model and its two different local environments.

### 3.1 Simplification of the robot's model and its motions

For our simulations, we assumed that the robot is a straight (inflexible), long, thin and weightless rod. Fig.4(a) shows -an inflexible section robot. If motor 1 rotates, the robot section moves in the *x0z* plane along the half-circle line marked blue in Fig.4(a).

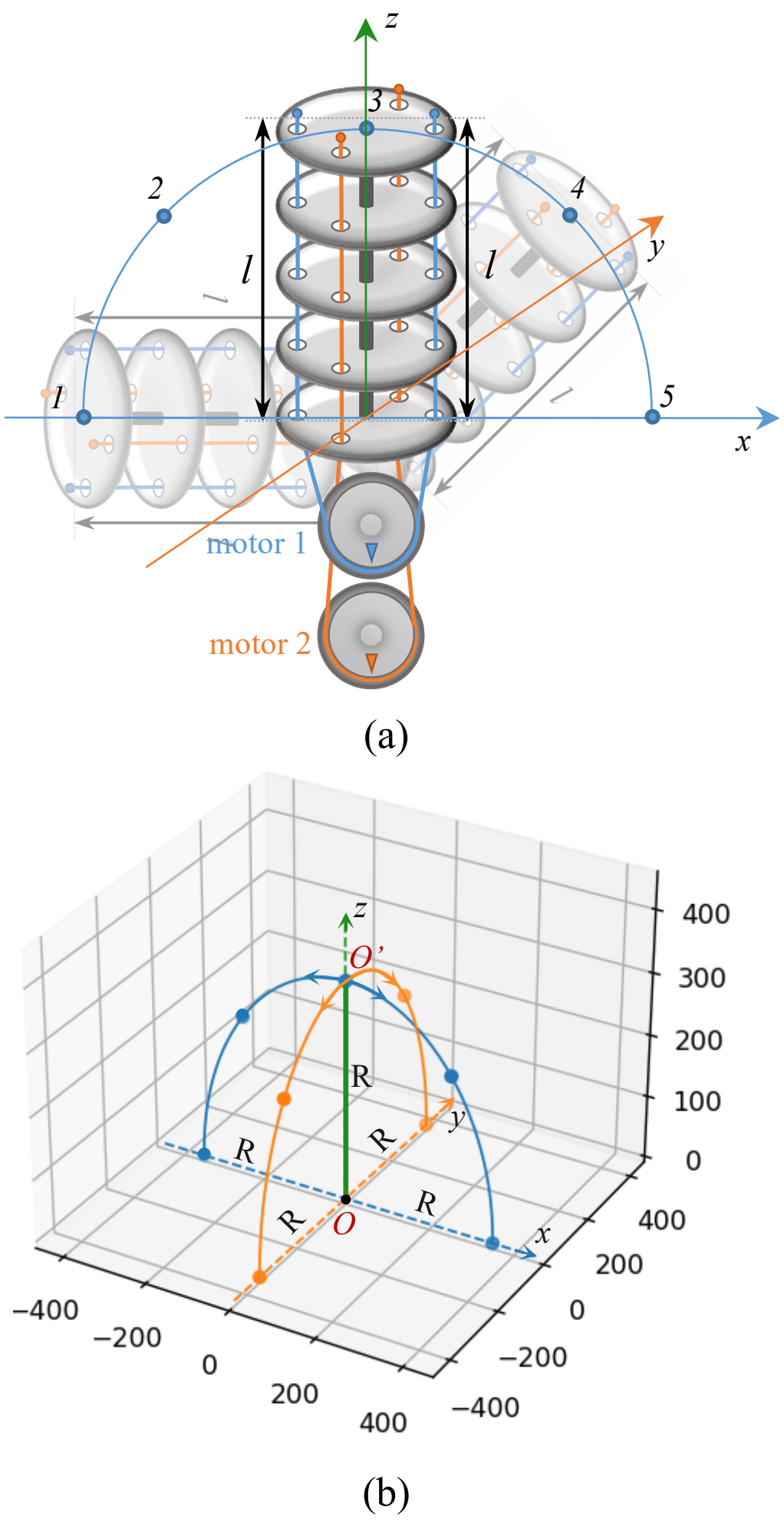

**Fig. 4.** Robot's section model: (a) – representation of a model which is similar to the physical prototype but not flexible; (b) – virtual (computer) model; the section is marked as a green solid line; it represents a straight (inflexible), long, thin and weightless rod. The blue arc represents the motion of the robot's section caused by the rotation of motor 1, the orange arc represents the motion of the robot's section caused by the rotation of the motor 2.

The robot's tip is located for example in points 1, 3 and 4. The lower disc is hinged at its centre, providing robot rotations in all planes orthogonal to the *x0y* plane. All the discs of the robot remain always parallel to each other. This model is an intermediate one between the physical prototype and the robot model in our simulations. Such an intermediate model will be necessary to explain the various criteria such as accuracy, motor damage, etc. Fig.4(b) shows the simulated model of the robot, i.e. it is a physical pendulum fixed at its lower point and performing motions caused by motor activity. In Fig.4(b), the robot's section is represented by a green solid line and initially located along the $z$ axis. The base point is $O$=(0, 0, 0). The tip of the robot's section is located at point $O'$=(0, 0, $R$). From this initial position the robot can move in 4 directions: left/right along the blue arc (i.e. along the blue marked half-circle in the $x0z$ plane) and forward/backward along the orange arc (i.e. along the orange marked half-circle in the $y0z$ plane). Please note that the projections of the blue and orange arcs on the $x0y$ plane are straight lines at the $x$ and $y$ axes with the maximum length 2$R$, i.e. [-$R$; $R$], respectively. In this paper, we use the term "main arcs" to denote these two arcs. Moving along the main arcs, each of the spatial coordinates $x$, $y$, $z$ of the robot's tip is changing in the range of {–$R$; $R$}. When the robot's tip is at any point in the environment, except for points on the $x0y$ plane, it can move in 4 directions, always along the arcs whose planes are parallel to the planes of the main arcs, as shown in Fig.5.

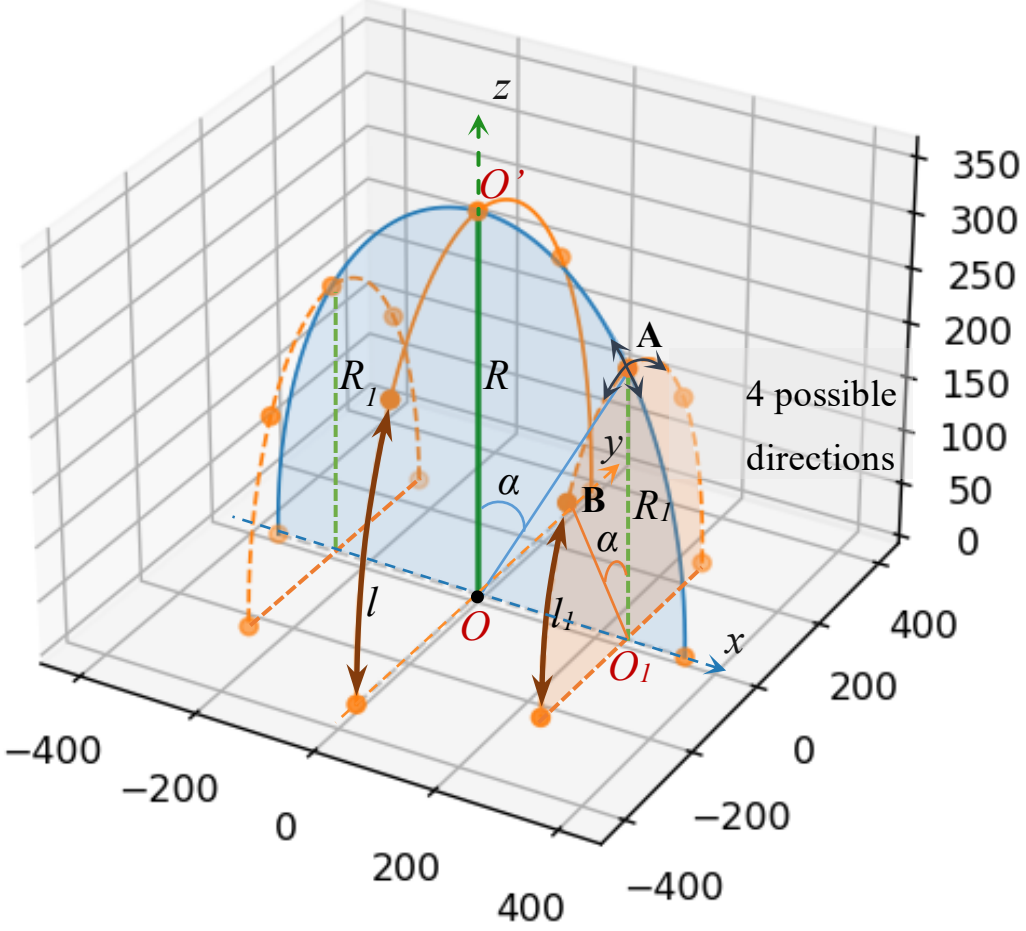

**Fig. 5.** The two main and two additional arcs, are shown with orange dashed lines

The plane of each new arc is parallel to the plane of one of the main arcs. The number of motor steps to pass the arc does not depend on the radius of the arc and is always the same i.e. $M$ motor steps. Thus, a single motor step moves the robot's section by an angle $\varphi = \pi/M$, and $m$ motor steps cause rotation of the robot's section $\alpha=m\pi/M$. For example, under the action of motor 1, the robot moves from $O'$ to $A$ along the blue main arc, see Fig.5. Then, under the action of motor 2, the robot's tip will move along the additional orange arc passing through point $A$. The plane of this additional arc (highlighted in orange) is parallel to the plane of the main orange arc. When the robot's tip reaches point $B$, the robot can either continue its movement along the arc $AB$ or move along a new additional arc whose plane is parallel to the main blue arc (see dashed blue arcs in Fig.6). Knowing the radius of the main arc $R$ and the angle $\alpha$ between the initial and the new positions of the robot's model, it is easy to calculate the spatial coordinates of point $A$ on the main blue arc, as well as the radiuses of the additional arcs and the spatial coordinates of the points belonging to the additional arc. The radius $R_1$ of the additional orange arcs in Fig.5 is $R_1 = R \cdot \cos(\alpha)$. The distance between point $A$ and $z$ axis is $R \cdot sin(\alpha)$, i.e. $A = (R \cdot sin(\alpha), 0, R \cdot cos(\alpha))$. Point $B$ has the spatial coordinate $B = (R \cdot sin(\alpha), - R \cdot cos(\alpha) \cdot sin(\alpha), R \cdot cos^2(\alpha))$. The same is true for the motion starting from the initial position along the main orange arc.

Since motor states change discretely, the robot's tip cannot stop at each point on the arc, but only at those that correspond to the motor state. Points in the environment which are the stop-points for the robot's tip can be represented with their spatial coordinates and with their motor states which we denote as motor coordinates. For example, point $O'$ has the spatial coordinates $(0, 0, R)$ and the motor coordinate $(0, 0)$. The robot can move to any point with motor coordinates $(i, j)$. where $i, j \in \{-M/2, -M/2 + 1, \dots, 0, 1, 2, \dots, M/2\}$. Please note that, from any point in the environment with the motor coordinates $(i, j)$, where $i, j \neq \pm M/2$, the robot can move in any of the four spatial directions by incrementing/decrementing each of the motor coordinates, and this causes a change of the spatial coordinates of the robot's tip. The example of the possible directions of motions for the robot's tip from the point $A$ is represented in Fig.5 with black arrows. The exceptions are the points where at least one of the motor coordinates is $i, j = \pm M/2$. From these points the robot can move only back to its previous stop-point. In order to illustrate the movement of the robot's tip with an example, we assume that the robot's motors execute $m = M/4$ steps with each activation. For this case, $\alpha = 45°$ and each arc possesses only 5 points. Due to the possibility to move from each stop-point into 4 directions, we obtain a sufficiently large set of points even for such a "coarse" environment discretisation, see Fig.6.

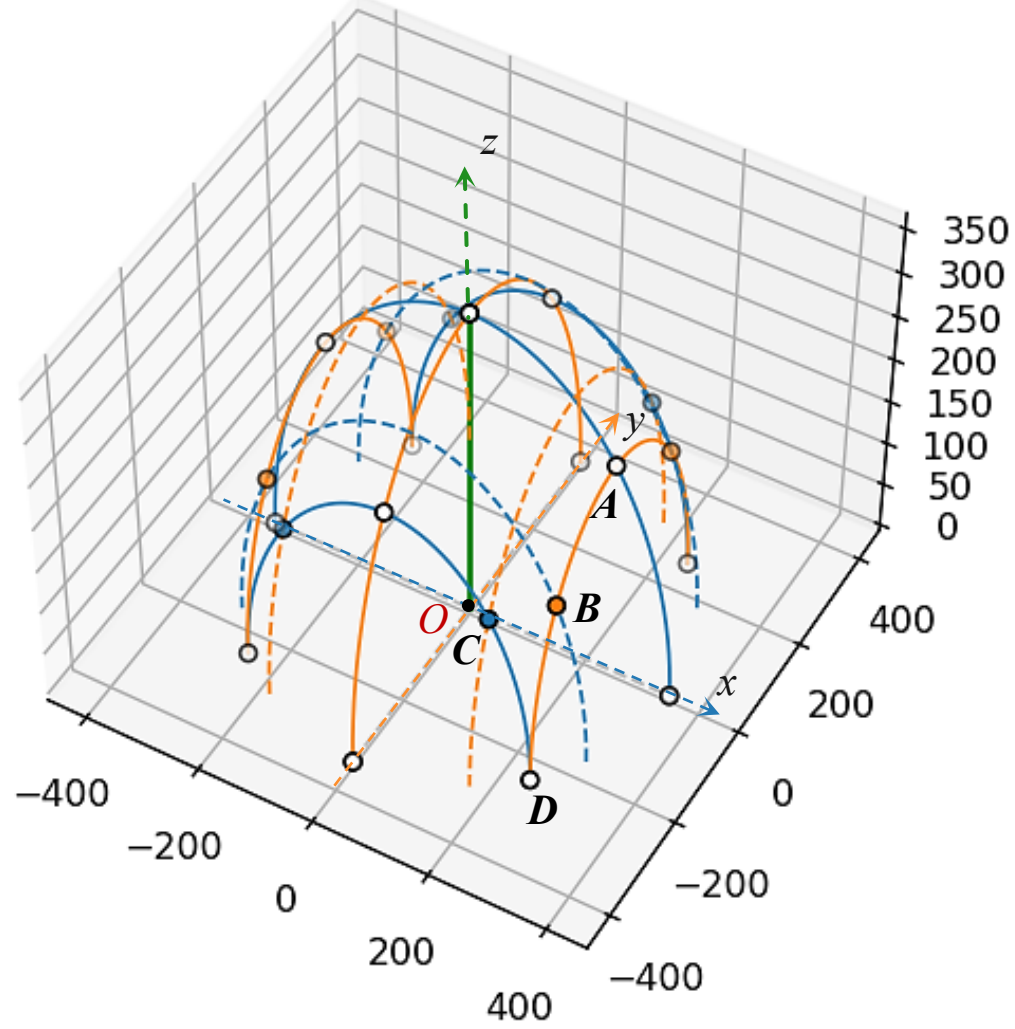

**Fig. 6.** Environment of the robot after its discretisation using $m = M/4$.

This environment contains 53 points with different spatial coordinates: 1 initial point O'; 4 points reachable from O' with 1 activation of a motor; 3 additional points reachable from each of previously mentioned 4 points that results into 3·4 =12 points; 5 additional points (2 in the one and 3 in the opposite direction) are reachable from 8 of 12 points resulting into 40 additional points. 4 of 40 points are the same. That results into 1+4+3·4+40-4=53 points with different spatial coordinates. However, the robot's model has only 25 different pairs of motor states, due to the fact that in our example the state of motor1 as well as motor2 can be represented with one of 5 possible values: -2*m*; -*m*, 0, *m*, 2*m*. Thus, some points with identical spatial coordinates correspond to different motor coordinates. For example, in Fig.6, two different motor coordinates (*m*, -2*m*) and (-2*m*, *m*) correspond to the same spatial point *D*. The first motor coordinate is achieved by following the path (0, 0)→(*m*, 0)→(*m*, –*m*)→(*m*, –2*m*), while the second motor coordinate is achieved by following the path (0, 0)→(0, –*m*)→(*m*, –*m*) →(2*m*, –*m*), i.e. the first path goes through point B, and the second path goes through point *C*. Additionally, the environment contains points with the same motor state, but different spatial coordinates. For example, points *B* and *C*. The path (0, 0)→(*m*, 0)→(*m*, –*m*) results in *B*=(*a*, –*b*, *c*), while the path (0, 0)→(0, –*m*)→(*m*, –*m*) results in *C*=(*b*, –*a*, *c*), with $a=R\cdot sin(\alpha)$, $b=R\cdot cos(\alpha)sin(\alpha)$, $c=R^2\cdot cos(\alpha)$.

To return from any point in the environment back to the initial point *O'* exactly, the motors should reproduce the reverse sequence of motor steps. Otherwise, by simply resetting the motor coordinates one by one (which is exactly what happens in the physical prototype, since the robot does not remember the perfect path), the robot's tip will arrive at a point close but not equal to *O'*, i.e. with some error. This leads to the importance/possibility of entering the accuracy criteria.

The example of the local environment consists of single-path (spatial) points and multipath points. A single-path spatial point can be reached only by a single path from the initial point *O'*, and a multipath point is a point reachable via multiple motion trajectories. For example point *C* is a single-path point because it can be reached only with the path (0, 0)→(0, –*m*)→(*m*, –*m*). Point *D* is a multipath point because it can be reached with two different motion trajectories, i.e. with the path (0, 0)→(0, –*m*)→(*m*, –*m*)→(2*m*, –*m*) as well as with the path (0, 0)→(*m*, 0)→(*m*, –*m*)→(*m*, –2*m*). Moving close to the *x0y* plane requires a large number of motor steps (meaning a significant energy consumption) even for small spatial movements. This fact can be taken into account in the distance and motor damage criteria, i.e., if more than 1 motion trajectory from the selected start point to the goal point exists, the set of criteria determining preferences by motions can be applied for the path selection.

### 3.2 Simplified local environments used in this work

For our experiments with the path planning algorithms, we implemented two simplified models of the robot's local environment. In the previous sub-section, we showed that even using a big step for the environment discretisation (*m*=*M*/4) we obtained 53 points taking into account all possible directions of the motion from each (spatial) point. The order of the environments, i.e. the number of its spatial points, as well as the presence/absence of multiple pathways should have a significant impact on how the path planning algorithms cope with a multiple-choice problem but also on the time required for the algorithms to solve the path planning task. To investigate these different aspects, we simulated 2 local environments with different properties. In Environment 1, the number of single-path points exceeds the number of multi-path points. Environment 2 consists of the multi-path points only. The virtual environments were programmed using the Python programming language and displayed using "matplotlib" library.

**Local Environment 1**

This model considers the fact that identical motor coordinates can correspond to different points in its environment. This model has single-path points, see Fig.7 (blue for the motor 1 and orange for the motor 2) and multi-path points (white points). In comparison to the simplified local environment with only 5 points per arc (see Fig.6) we used a model with 11 points per arc but we considered only the additional arcs formed from the main arcs. This means that only two stage moves are considered i.e. the first one along the main arc and the second one along one of the additional arcs. The 11 points per arc result in a discretisation of m=70 real motor steps.

Fig.7 shows Environment 1. It has two main arcs (blue and orange solid lines) and additional arcs (grey dashed lines). This configuration has 165 points in total: 64 single-path points that can be reached by using motor 1 only (blue points), 64 single-path points that can be reached by using by motor 2 only (orange points) and 37 multi-path points which can be reached by using both motors (white points).

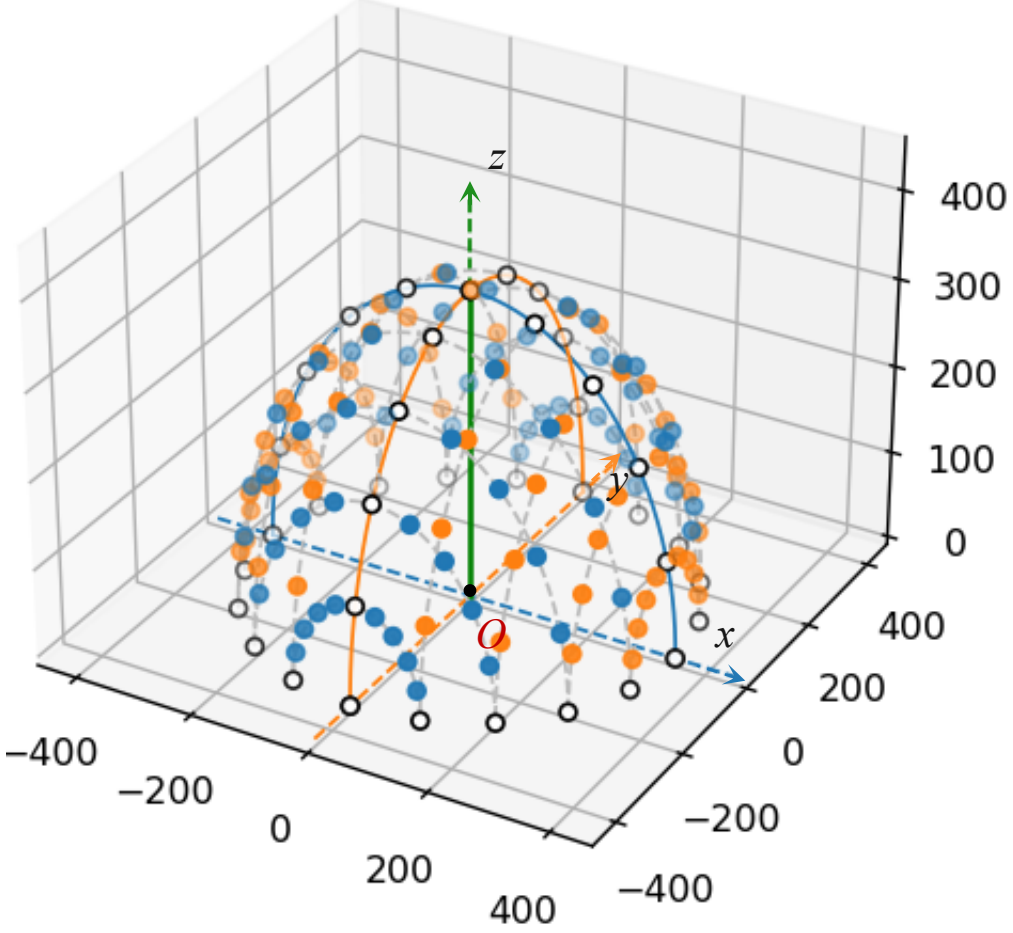

**Fig. 7.** Local Environment 1 with single- and multi-path points.

**Local Environment 2**

This model is a more abstracted and simplified environment, in which the actual features of robot motion were not really taken into account. Each point in this environment described by motor coordinates corresponds to one and only one point in space i.e. a one-to-one mapping from the motor coordinates to the spatial coordinates, see Fig.8.

This environment consists of 61 points in total. It consists of 11 arcs parallel to the *x0z* plane and 11 arcs parallel to the *y0z* plane. In contrast to the local Environment 1, each arc has a different number of points. The distribution of points between arcs can be described as 1-3-5-7-9-11-9-7-5-3-1. This distribution pattern is the same for arcs in both planes. In this model we consider that arcs are intercepting. The points of interception are marked white.

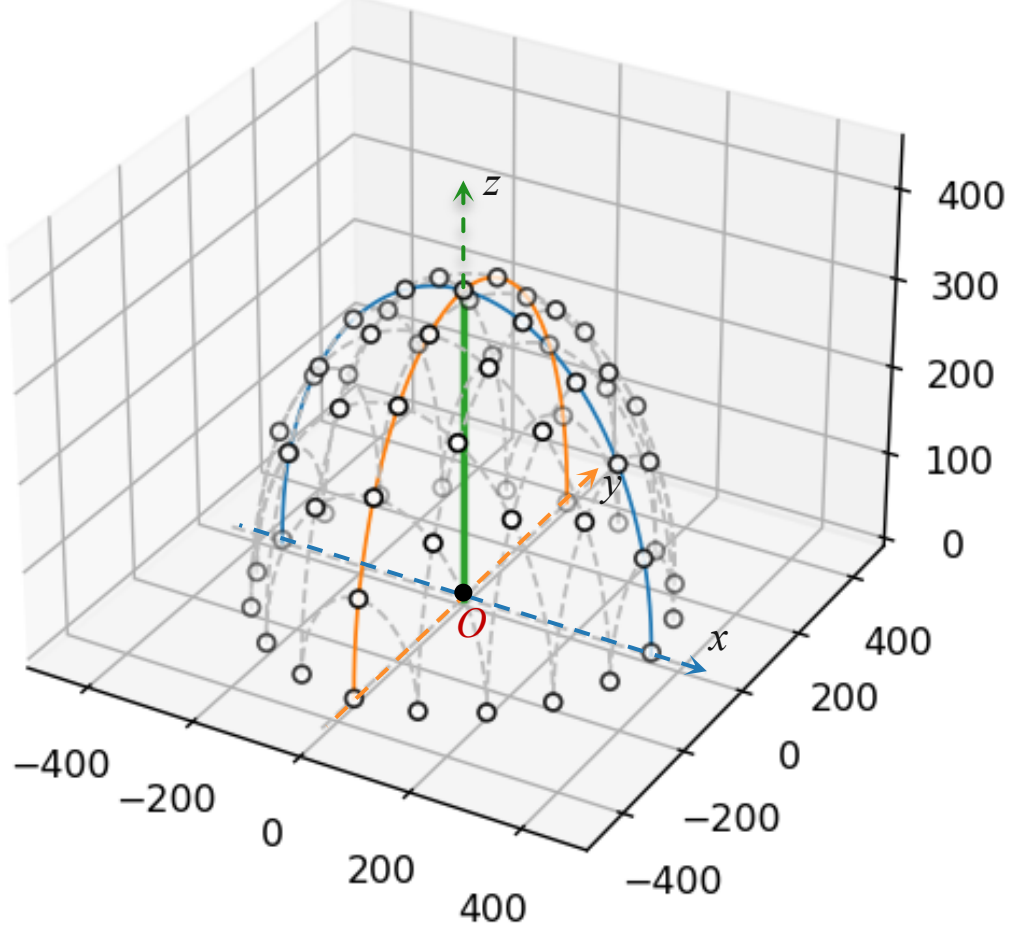

**Fig. 8.** Local Environment 2 with multi-path points only.

### 3.3 Modelling Global Environments

Environment 1 and Environment 2 are local environments. Fig.9 illustrates how we calculated the global environment of the 2-sections robot's model on the example of local Environment 2 for each section of the robot. Fig.9(a) the lower robot's section is marked blue and its local environment is the set of green points. The tip of the lower (blue) section is located at point Q, the base point of the upper (orange) section. The upper section is marked orange and its environment is the set of red points. The tip of the upper section is located at point P.

Instead of representing each point of the local environment with its spatial coordinates, we define that each point in the local environment can be uniquely identified by an index. The indices of the points in the local environment are integer values starting from 0 to $N$-1. In Fig.9(a) the point Q, which belongs to the local environment of section 1 (see set of green points), has index 48. Point P, which belongs to the local environment of section 2 (see set of red points) has index 12. Therefore, if the point P is considered as a point of the global environment, it can be represented by two indices as follows e.g. {48; 12}. This representation clearly defines the position of the robot in its global environment. For each new position of the lower section of the robot, the local environment of the upper section is a new set consisting of $N$ points. Thus, the global environment of the 2-sections model consists of $N^2$ points. The global environment constructed using local Environment 2 for each section of the robot consists of 612=3721 points and is shown in Fig.9(b). The global environment of a 2-section model constructed using the local Environment 1 consists of significantly more points, i.e. 1652=27225 points.

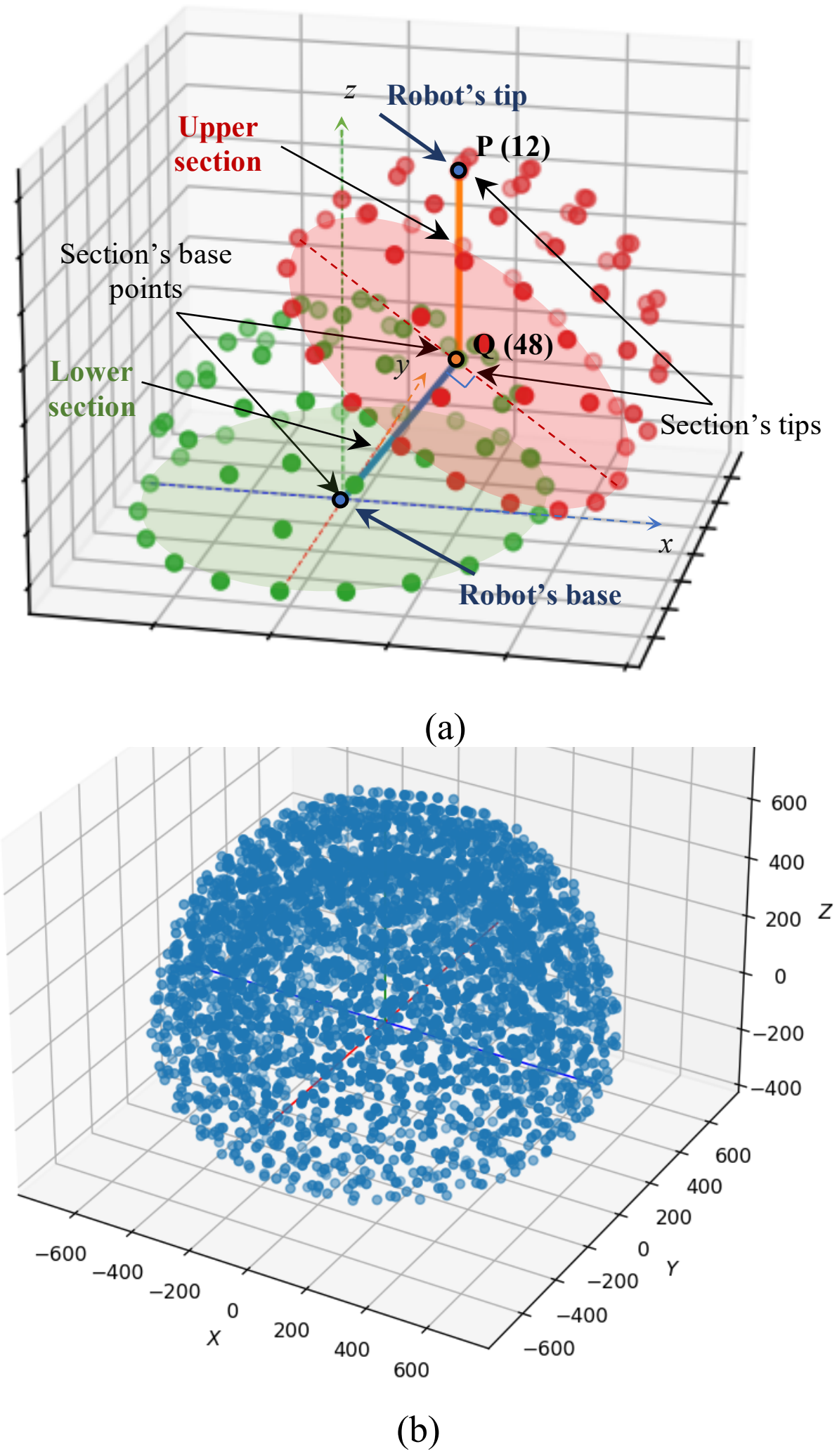

**Fig. 9.** Global environment of the 2-sections robot simulated using local Environment 2 (3721 points in total).

The cardinality of the global environment of a multi-section robot model increases exponentially, depending on the number of sections. For a 4-section model of the robot, a point *G* in its global environment can be represented by a set of 4 indices {*N1*, *N2*, *N3*, *N4*}, where each index *Ni* defines the position of the i$^{th}$ section's tip in its local environment. Moreover, point G can be represented by different sets of indices {*NN1*, *NN2*, *NN3*, *N4*}, due to the fact that the set determines not only the position of the robot's tip but also the shape of the robot, and a multi-sectional robot can change its shape for a fixed pair of points $O = (0, 0, 0)$ and $G$ due to its increased degrees of freedom. We denote the representation of the points in the global environment using the set of their indices as "index coordinates".

## 4 Algorithms explored

The current state-of-the-art in motion planning for continuum robots is based on the use of motion control models. One of the most popular motion control model is the Cosserat theory [11]. Many researchers [5–9] utilize it for modelling motion control, which considers different kinds of deformations of physical objects, but requires a long time and significant computation costs. A discrete Cosserat approach, which treats the continuum robot prototype as a finite set of constant-length pieces, is suggested in [7]. This approach transforms the problem into a discrete one, making its solution easier and reducing computation costs. [12] provides a comparison of different motion models and demonstrates that the Cosserat theory can be used for the tasks where high accuracy is needed, but it is not suitable for real-time applications. The so-called model-free approaches [13–15] help to solve the motion problem of continuum robots faster, as they do not consider mathematical background of continuum robots' motions. In our research, we aim to find a solution that combines suitable continuum robot path planning algorithms with a decision-making algorithm to generate paths based on the environmental state and the robot's state.

There are few traditional algorithms for path planning for mobile robots. They include A* algorithm, Genetic algorithm, Ant Colony Optimisation algorithm and Rapidly Exploring Random Tree algorithm [16, 17]. In accordance with our definition of autonomy, the robots are required to determine the optimal path based on unexpected changes in the environment or internal state parameters. Given the focus of our research interests on decision-making algorithms, we were looking for existing solutions that are combing path-planning and decision-making.

Among many well-known decision-making algorithms [18–20], AHP is the most popular one, it is well documented, easy to use and can also be used for path planning tasks. For this purpose the AHP algorithm is usually combined with the dynamic programming A* algorithm or A*-based algorithms, as it is the most common path planning algorithm that works fast and does not need to be optimized [21–24].

Looking for the second path planning algorithm candidate, to have some comparison, we selected the Genetic algorithm. This algorithm can be easily adjusted to the multiple-criteria decision-making, as shown in [25], where GA is used together with AHP for tourism route planning. It was also used for continuum robots path planning [26], however authors did not use any decision-making algorithm. The Ant Colony and other flavours of Particle Swarm Optimization (PSO) algorithms tend to be faster than GA. Nevertheless, we selected the GA over the PSO due to the fact that PSO, in contrast to the GA, relies on a continuity of solution space that is not always the case for continuum robots. This means that not all closely spaced trajectories are realizable, and not all destinations may be reachable from different directions, especially in a changing environment.

Even though there are many already existing solutions and reports, we did not find researches that used these algorithms for the path planning for continuum robots using decision-making algorithms. The most of researches mentioned above conducted experiments using mobile robots such as small cars, or just solve the path generation problem in general and did not apply it to any kind of real device such as robot/vehicle, etc.

We decided to choose both, A* algorithm and GA, and modify these algorithms, so they can be used with AHP to suit our experiments. We made changes to the classical representations of both algorithms primarily by adding multiple criteria for assessing the generated paths.

This section describes the algorithms used in our experiment, which aims at comparing the two path planning algorithms: GA and A* algorithm, when more criteria than just the path length have to be considered. The criteria weights were generated using the AHP decision-making method.

### 4.1 Analytical Hierarchy Process

Multiple criteria decision-making is a process that identifies the optimal alternative from a set of feasible alternatives. In multiple criteria decision-making, a problem with *m* alternatives and *n* criteria can be expressed in a matrix format as shown in Eq. (1):

$$M = \begin{pmatrix} A_1 \\ A_2 \\ \vdots \\ A_m \end{pmatrix} \begin{matrix} w_1 & w_2 & \cdots & w_n \\ C_1 & C_2 & \cdots & C_n \\ \end{matrix} \begin{pmatrix} z_{11} & z_{12} & \cdots & z_{1n} \\ z_{21} & z_{22} & \cdots & z_{2n} \\ \vdots & \vdots & \cdots & \vdots \\ z_{m1} & z_{m2} & \cdots & z_{mn} \end{pmatrix}, \quad (1)$$

where $A_1, A_2, \ldots A_m$ are feasible alternatives to select the decision from, $C_1, C_2, \ldots, C_n$ are evaluation criteria, $z_{ij}$ is the given performance rating of alternative $A_i$ under criterion $C_j$, and $w_j$ is the weight of criterion $C_j$, which is determined based on given criteria importance [27].

AHP is the most common multiple criteria decision-making algorithm developed by T. L. Saaty in 1980 [28]. It is applied in various fields such as business, marketing, risk analysis for safety enhancement, environmental location selection, performance evaluation, and path planning. The algorithm is also applied in some online open-source tools [29–31], AHP has a hierarchical structure (see Fig.10) and makes decisions based on pairwise comparisons of alternatives or criteria. The AHP hierarchy contains three levels: goal (decision), objectives (criteria), and alternatives (potential results). The AHP is straightforward to use and employs a scale of different importance levels (see Table 1) to perform pairwise comparison of objectives and estimate weights [21, 24, 25]. The importance levels should be selected manually.

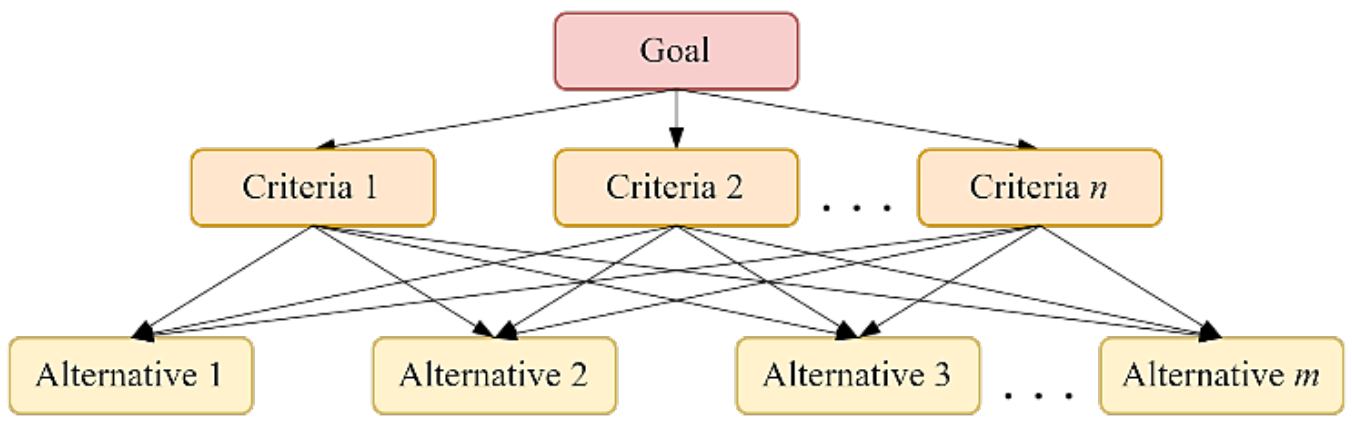

**Fig. 10.** AHP hierarchical structure

A full description of the algorithm can be found in [21]. Here we give the sequence of main steps of the algorithm.

1. Problem analysis;
2. Generation of the hierarchy structure (see Fig.10);
3. Definition of the relative importance of the criteria using a pairwise comparison scale;
4. Calculation of criteria weights;
5. Evaluation of the consistency indexes for criteria;
6. Evaluation of alternatives with reference to each criterion.

Let us take a closer look to the step 3 of the AHP algorithm, as it is a crucial step that affects the weights calculation the most. AHP uses a pairwise comparison technique to set the relation/priority between criteria. This means, that all possible pairs of criteria have to be observed, and criteria within each pair have to be compared determining which criterion from the pair has a higher priority over the other one. To do so, a user should use a pairwise comparison scale (see Table 1) and manually assign relative importance values to each of the pairs, based on the knowledge about the criteria and the task. The relative importance of the criteria (further relative importance) is a value, which represents the relationship between criteria and thus allows to determine their relative priority, i.e. a criterion is of greater or lesser importance. The pairwise comparison scale (i.e. Table 1) is provided by the AHP algorithm authors and remains constant.

There is a small example of how the relative importance values are assigned to the criteria. Assume that there is a task with three different criteria $C_1$, $C_2$ and $C_3$. The criteria should be combined in the following three different pairs[2]: ($C_1$, $C_2$), ($C_1$, $C_3$), ($C_2$, $C_3$). The value of relative importance for criteria in each pair are assigned by user. Assume that the user chooses the relative importance value 3 (moderate importance) for the pair ($C_1$, $C_2$), value 5 (strong importance) for ($C_1$, $C_3$) and value 9 (extreme importance) for ($C_2$, $C_3$). Then, under these conditions we can say that $C_1$ is three times as important as $C_2$, $C_1$ is five times as important as $C_3$, and $C_2$ is nine times more important than $C_3$. This means also that the values of relative corrections for the pair with the permuted criteria are 1/3 for ($C_2$, $C_1$), 1/5 for ($C_3$, $C_1$) and 1/9 for ($C_3$, $C_2$). Relative corrections of a single criterion versus other criteria are added to obtain the assigned relative importance for this criterion. Based on the assigned relative importance values, the normalized weight for each criterion can be calculated as described in [21]: $w_1$ = 0.589 for criterion $C_1$, $w_2$ = 0.344 for $C_2$ and $w_3$ = 0.067 for $C_3$.

[2] To form groups of criteria, combination without repetition is used.

The output of the AHP algorithm is a numerical value of a weight ($w$), one for each of the defined criteria, where $0 < w \leq 1$. A description of the weights calculation can be found in [21]. The weights' calculation can be done using one of the online open source tools e.g. [29].

**Table 1.** Definitions and values of relative importance of criteria Ci, Cj based on their comparison

| Relative Importance Value | Definition of relative importance |
|---|---|
| 1 | Equal importance of both criteria |
| 3 | Moderate importance of $C_i$ over $C_j$ |
| 5 | Strong importance of $C_i$ over $C_j$ |
| 7 | Very strong importance of $C_i$ over $C_j$ |
| 9 | Extreme importance of $C_i$ over $C_j$ |
| 2, 4, 6, 8 | Intermediate values between two adjacent judgments |

### 4.2 Application of AHP for Our Task: Multi-Criteria Decision-Making

To use the AHP decision-making algorithm we established four criteria that are crucial for enhancing the resilience of a continuum robot. This includes reducing or avoiding mechanical damage and increasing time to maintenance. Thus, we used the following four criteria: distance, robot's tip accuracy, motor damage, and mechanical damage.

*Distance*: estimates the length of the generated path. This criterion is important for the path planning task. Prioritising this criterion will force the robot to choose the shortest path among all other alternatives. Following the shortest path will save energy and reduce the time the robot moves, which are two of the most important resilience features. All distance calculations mentioned further in this paper are made using Euclidean distance [32].

*Motor damage*: estimates the damage that will be caused to the motors by following the generated path. This criterion aims to prolong the time to maintenance of the setup and particularly of the stepper motors. Prioritising this criterion means that the path will be generated based on the state of the stepper motors, i.e. how many steps each motor did since the beginning of its work. This ensures that all motors will be used evenly.

*Mechanical damage*: estimates the damage will be caused to robot's wire cables after the robot follows the generated path. As it is explained in section 2 and shown in Fig.1 the cables run along the entire length of the robot. Under the stress generated by the motors, the cables cause the robot to bend in a given direction, leading to damage in the place of bending. Prioritising this criterion makes the robot to find the path to the goal point that will cause the least possible mechanical damage.

*Robot's tip accuracy*: estimates how accurate the robot's tip reached the goal point. This criterion helps to generate the most precise path with minimum tip error, which we determine as the distance between the coordinates of the robot's tip[3] and the coordinates of the goal point. The less the distance, the lower the tip error, and the higher the accuracy. Generating a path with high accuracy is important for tasks that require precise movements. This criterion is influenced primarily by the results of the search algorithm for alternative goal points (section 5.1), which search for the alternative goal points.

Table 2 gives a short overview of the criteria described above.

**Table 2.** criteria for decision-making in Analytical Hierarchy Process

| # | Criteria name | Description | Resilience properties |
|---|---|---|---|
| 1 | distance | estimates the length of the generated path | energy and time savings |
| 2 | motor damage | estimates damage of motors depending on the path selected | extended time to maintenance of the motors |
| 3 | mechanical damage | estimates damage that will be caused to robot tendons after the robot followed the generated path | extended time to maintenance of robot wire cables |
| 4 | robot's tip accuracy | estimates how accurately the robot's tip is at the goal point | accurate work |

The established criteria are used to evaluate the paths, generated by the path planning algorithms. In order to achieve this, four evaluation functions were created one for each of the criteria:

- $f_{dist}$ – calculates the path distance. The total distance of the whole path is a sum of distances between each two neighbour points in the path. To calculate the distance the space coordinates of the points are used.
- $f_{mot}$ – calculates how many steps the motors will do to follow the path. To calculate the motor damage that a path causes to the robot, we calculate the number of steps each motor did to follow this path. The number of steps that each motor did during the whole time of its work, is stored in a *sqlite3* database. These values are increased after each movement.
- $f_{mech}$ – calculates the mechanical damage that will be caused to the robot wires. The "mechanical damage" criterion aims to prevent robot to repeatedly go through the same segments of the environment. An *sqlite3* database stores all segments of the environments the robot's tip ever passed as well as the number of times the robot did it (mechanical damage of this segment). We define, the "segment of environment" as two neighbour points taken from the generated path. The mechanical damage is calculated for each two neighbour points in the path. The values in the database are updated after each movement.
- $f_{acc}$ – calculates the accuracy of the robot's tip. To calculate the accuracy of the path the distance between the space coordinates of the robot's tip and the space coordinates of the goal point is calculated.

[3] The robot's tip is considered as a mathematical point in our experiments, i.e. the robot's tip has no length, width, or height. In other words, it has no size.

These functions are also called *fitness functions*. They are means to evaluate the quality of generated path. To make these functions sensitive to the criteria and the priorities of the criteria, a *multi-fitness functions* is used in this work. The multi-fitness function represents the sum of the weighted fitness functions of each criterion. The result of the multi-fitness function is a numerical value (further *fitness value*). The smaller the fitness value, the better the path. The multi-fitness function we use in this paper is shown in Eq. (2), its more general representation is given in Eq. (3).

$$F = w_{dist}f_{dist} + w_{mot}f_{mot} + w_{mech}f_{mech} + w_{acc}f_{acc} \quad (2)$$

$$F = \sum_{i=1}^{k=4} w_i \cdot f_i \,, \quad (3)$$

where $F$ refers to the fitness value for the whole path, $i$ refers to the criteria from Table 2, $k$ refers to the maximum number of criteria we used in the experiment, $w_i$ refers to the criterion weight obtained by the AHP algorithm and $f_i$ refers to the criterion's fitness function.

A correctly composed fitness function increases the algorithm's output quality. It is tailored and adjusted for each use case to accurately reflect the algorithm's purpose.

### 4.3 Genetic Algorithm

Genetic algorithm (GA) is a global search and optimisation technique modelled on natural selection, genetics, and evolution. The algorithm is inspired by Darwin's theory and was first introduced by J.H. Holland in [33]. GA can be applied to a wide range of disciplines, such as mathematics, medicine, and engineering [34].

GA has a clear structure. First, a population of individuals (chromosomes) is generated. Each chromosome consists of a sequence of *genes* and represents a solution to the problem. In our case the solution is a path represented as a sequence of points, from the start to the goal points, where each gene is represented by the index of the point (see section 3.3). The first gene in the chromosome represents a start point, the last gene is a goal point. If the start point is equal to the goal point, the chromosome consists only of one gene. The points in the chromosome between the start and goal points represent the points of the path that the robot should follow. The size of the population remains constant during the whole process of the algorithm's work. GA modifies the chromosomes in the population using genetic operators such as *evaluation, selection, crossover*, and *mutation* until the algorithm reaches a terminal state. The terminal state can be defined as either maximum number of *generations* which is the number of algorithm iterations or as a specific fitness value of the obtained solution [35]. Once the terminal state is reached, the best chromosome from the population is selected as the result of the algorithm's work. The conditions determining which chromosomes are good are set by the developers of the algorithm, for example, to choose a chromosome with the lowest fitness value according to the definition of the multi-fitness function. The population size and number of generations affect the population diversity and how the chromosomes in the population are going to be modified during generations. After the terminal condition is reached, the chromosome with the lowest fitness value from the population is selected.

Fig.11 shows the structure of GA with genetic operators selected for our investigations. In the rest of this sub-section we describe each step of GA in detail and discuss the approaches that were selected for the genetic operations of the algorithm.

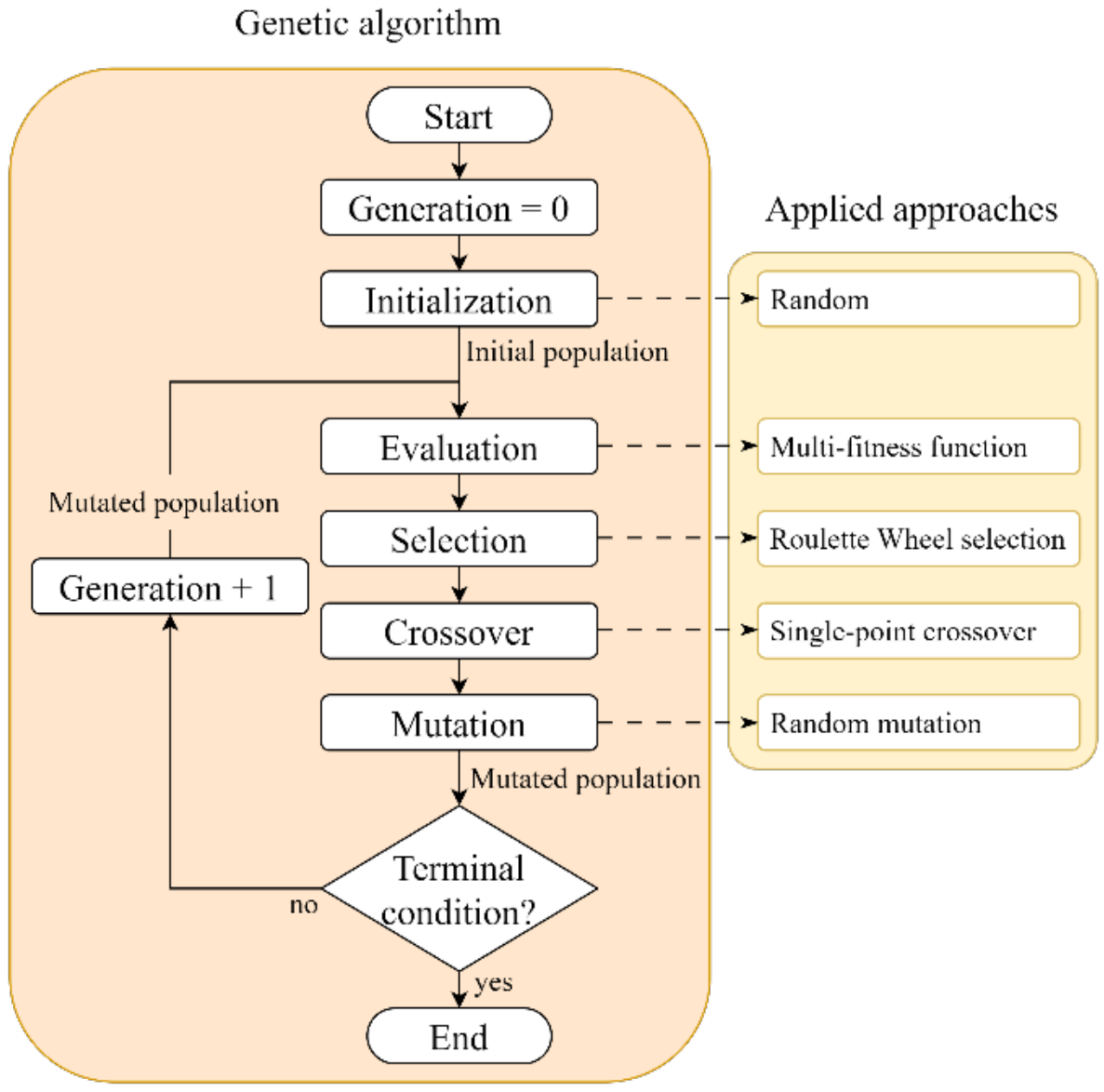

**Fig. 11.** Steps of Genetic algorithm and approaches applied in each step in our experiment

**Initialisation of Chromosomes**

The first and most important step of the entire GA is the creation of an initial population, or initialisation of the population. There are some requirements to the initial population. The population has a fixed size and consists of chromosomes, each of which consists of genes. The chromosome length is not necessarily fixed. During the initialisation step, genes are generated and combined into chromosomes until the whole population is ready. The chromosomes in one population are not necessarily all unique, they can repeat themselves. Fig.12 shows the population structure schematically.

**Fig. 12.** Population, chromosomes and genes

The example population in Fig.12 consists of 5 chromosomes, i.e. the population size is equal to 5. Each chromosome consists of genes. Each gene corresponds to a single point, which is represented by its index coordinate in a local environment of a robot's section. The index of the start point is 8, the index of the goal point is 4. The start and goal point are the same for each chromosome in the population and do not change during generations. The five chromosomes represent five different solutions, i.e. 5 different paths, each of these solutions is going to be processed by the genetic operations until the terminal condition is reached.

The most common approach to initialize a population for GA is a random initialisation. Chromosomes are filled with a set of genes that are randomly generated or picked from the set of existing points. This approach provides a diverse population, but it can significantly impact the algorithm's performance. The next popular approach is Latin Hypercube [36]. It is a sampling method used to generate a near-random sample of parameter values, that provides more optimal solutions than pure random approach.

In our implementation we used an improved random approach that is adjusted especially for our problem and uses the specifics of points' position in the robot's environment. A list of possible connections between points, which the robot can reach, was created and used for the initialisation. The starting point is always known, as it equals the last tip position of the robot after the previous motion. Each next point $n_i$ is randomly chosen among possible connections of the previous point $n_{i-1}$. This approach still initializes random chromosomes, however, in comparison to a completely random approach, it uses only neighbour points to create a point sequence, and it also avoids dead ends. This helps prevent the robot from moving "in circles". The chromosome initialization process ends if either the maximum chromosome length is reached or if during chromosome generation, the algorithm selects the goal point as a gene. These features provide more meaningful results and increase the efficiency of GA, in terms of fitness values of the resulting paths.

**Evaluation of Chromosomes**

Once the population has been initialised, the GA begins to perform genetic operators one by one, starting with evaluation of chromosomes. Each chromosome in the population is evaluated individually using the multi-fitness function $F$ from Eq. (3) and obtains its own fitness value. The fitness value of a chromosome indicates how well the chromosome fits the current solution. For the population of chromosomes, the chromosome with the lowest fitness value is considered the best.

**Selection of chromosomes**

During the selection step, the algorithm chooses parent chromosomes for crossover and mutation. The population after chromosome modifications is denoted as offspring population.

There are several algorithms for selection of parent chromosomes. The most popular are Roulette Wheel Selection, Elitism Selection, Tournament Selection, Stochastic Universal Sampling, Linear Rank Selection, Exponential Rank Selection, and the Truncation Selection [37]. Selection algorithms can be basically classified into the two main types: deterministic and stochastic algorithms. Deterministic algorithms always choose parent chromosomes with the best fitness value, but in some cases, they can cause the population to stop evolving. Stochastic algorithms choose parent chromosomes randomly. However, in this case, the method does not take fitness values into account. Roulette Wheel Selection is one of the most often used algorithm that stands between these two approaches. This is a stochastic selection method, in which the probability of selecting of an individual is proportional to its fitness values. In other words, the better the chromosome's fitness value, the more likely the chromosome is to be selected [38]. In our implementation of GA, we use the Roulette Wheel Selection algorithm as it combines the advantages of both deterministic and stochastic algorithms. Some parent chromosomes are chosen randomly, another based on their fitness value. The size of the resulting population of parent chromosomes is equal to the initial one, as due to the Roulette Wheel Selection algorithm some chromosomes are selected multiple times, and some were not select at all.

**Crossover operator**

The crossover operator (further "crossover") modifies chromosomes in the population to create a diversity for future generations. Crossover operates on parent chromosomes, using their genes to create new offspring chromosomes. Common types of crossover include single-point, two-point, k-point, partially mapped, and shuffle crossover [37].

In our implementation we use a single-point crossover. In this crossover type the parent population is segmented in groups of two, where each pair of two parent chromosomes is used to produce two offspring chromosomes and as result create an offspring population. A single-point crossover uses a separator, which position is randomly chosen. The separator divides each parent into two parts. The first parts of both parent chromosomes remain unchanged, while the second parts are swapped between each other to create the offspring. Fig.13 shows an example of a single-point crossover. In the example each chromosome consists of 6 gens. We set the separator after the 2nd gene of each parent. The parts of the parent's chromosomes before the separator are coloured black and those after the separator are coloured green and red, respectively. The green and red parts are swapped creating offspring chromosomes.

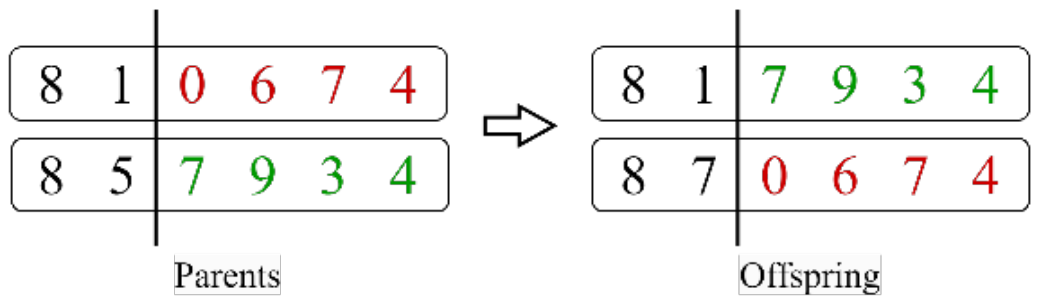

**Fig. 13.** Single-point crossover

**Mutation operator**

At the mutation step, only a small part of a parent chromosome is changed. In contrast to a crossover, where parent chromosomes are used in pairs, mutation operates with single chromosomes. The most common mutation operators include random mutation, swap mutation, scramble mutation, inversion mutation [39].

In our implementation we use a random mutation. A single gene of the parent chromosome is changed to another randomly generated value (see Fig.14). The parent chromosome to which the mutation operator is applied, is also randomly selected, so only a few chromosomes in the entire population are mutated. Whether or not a chromosome is mutated depends on the mutation rate value, which is set by the user before the algorithm begins to work.

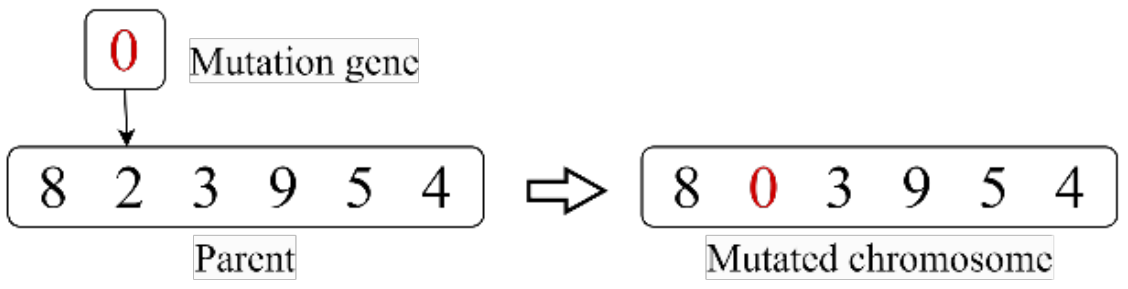

**Fig. 14.** Random mutation

**End of the algorithm – algorithm results**

After 3 generations (algorithm repetitions), the algorithm provides a final population that contains 50 paths (chromosomes). The path with the lowest fitness value is considered as the resulting path.

## 4.4 Improved Genetic Algorithm

This sub-section contains information about our improved GA. Since the GA is quite time-consuming, some optimizing measures were taken to improve the run time of the algorithm. The weak point of the GA is that each step will be implemented in the loop for each chromosome in the population. Thus, the run time of the algorithm directly depends on the population size (the larger the population the longer the run time). In order to address this issue, we propose an improved model of GA in this work (see Fig.15).

In the improved model, we reduced the number of iterations the algorithm does, to decrease the execution time. It was achieved due to the changed structure of the algorithm. In the classical GA (see Fig.11), each chromosome in the population is evaluated in each generation. Even if a chromosome remained unchanged during the processing it will be evaluated again. But this is an unnecessary step as the same chromosomes have the same fitness values, and it should not be calculated again. We also noticed that often, especially when the algorithm should generate short paths, the population contains same chromosomes, therefore, it is redundant to calculate the fitness value of each chromosome in the population, as the values remain unchanged. In our improved version of the algorithm (see Fig.15) we propose to evaluate only unique chromosomes.

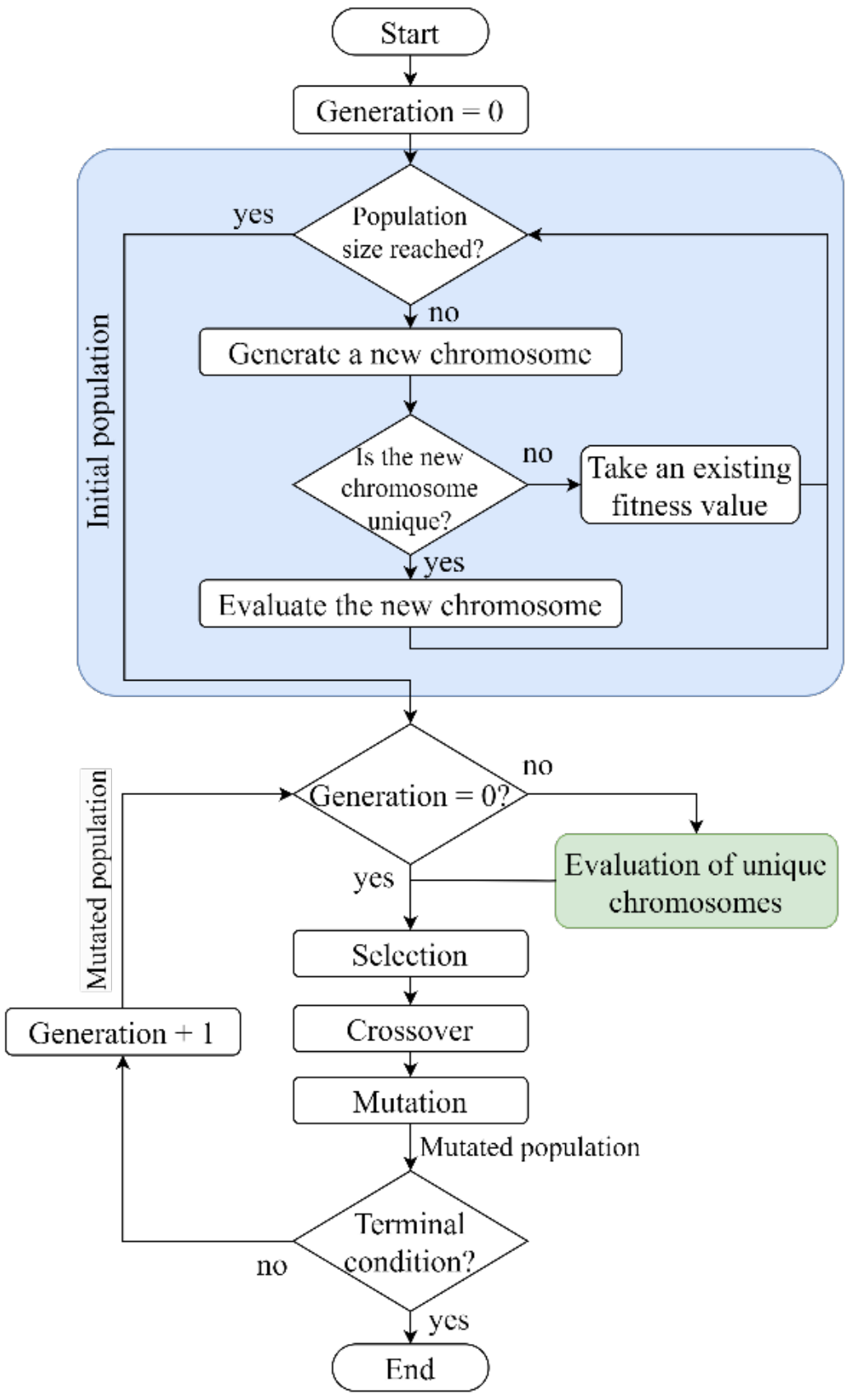

**Fig. 15.** Improved Genetic algorithm

First, we combined initialization and evaluation steps to avoid one extra loop (see blue area in Fig.15). The algorithm first generates a chromosome, and checks if it is unique i.e. if it appears in a population for the first time. For a unique chromosome the evaluation multi-fitness function is performed, for a repeating chromosome, the algorithm just takes an existing fitness value, without calculating it again. This combination of initialisation and evaluation is applied only once at the very beginning, when the generation number is equal to zero.

For mutated populations (i.e. if Generation > 0) the algorithm keeps evaluating only unique chromosomes (see green block in Fig.15). Thus, our improvement allows all chromosomes in each generation to be evaluated without unnecessary repetitive calculations.

The important aspect is that same chromosomes are having the same fitness value, only within the same task and by using the same fitness function. Using a different fitness function will lead to the different fitness values even for the same chromosomes.

During the experiments, we also noticed that there is a high chance of getting invalid chromosomes after performing crossover or/and mutations. An invalid chromosome is a chromosome that does not contain any neighbouring points, i.e. there is no direct connection between points $n_i$ and $n_i$+1. For this reason, the fitness function includes an additional check that verifies that adjacent points on a chromosome are directly connected. If not, the chromosome is considered invalid and its fitness value is increased, reducing its chances of being selected in the next generation.

The improved algorithm represented in Fig.15 uses the same approaches for the genetic operators that were described in the section 4.3 and shown in Fig.11.

### 4.5 A* Algorithm

The A* algorithm is a well-known path planning algorithm. It was proposed by Peter Hart, Nils Nilsson, and Bertram Raphael [40]. It is widely used in various domains, such as game development [41], logistics and path planning for mobile robots [22, 42–44].

The core idea of the algorithm is to find the minimum distance between the starting and the goal points. A* achieves this by using heuristic search. It calculates the cost of moving from the current point n to its neighbour points, and creates a sequence based on the cost results. The point with the lowest cost is given a higher position in the sequence. The cost function is presented below:

$$f(n) = g(n) + h(n) \quad , \tag{4}$$

where *g(n)* denotes the cost from the starting point to the current point *n* and *h(n)* denotes the cost from the current point *n* to the goal point.

The classical A* algorithm described in [40] involves the following steps described in Fig.16. In Fig.16, “open list” refers to a list of points that are yet evaluated, while “closed list” refers to a list of points that have already been evaluated. The cost function *f(n)* calculates the distance between points using either Euclidean or Manhattan distance [32]. Based on the result of the cost function A* provides a path that is considered to be the best under the current environmental conditions.

### 4.6 Improved A* Algorithm

The classical A* (see Fig.16) takes only the only distance between points into account, while we are interested in multi-criteria assessment. Thus, in our experiments we kept most of the steps of the classical approach but changed its cost function *f(n)*. This means that instead of cost function *f(n)* from Eq. (4) we used the multi-fitness function *F* from Eq. (3), i.e. the same that was used for GA. It allows generating the path taking into account multiple criteria, and also helps to have a clear output comparison and analysis of both path planning algorithms.

Unlike GA, where the calculation of a fitness value is applied to the entire path, the classical A* algorithm can only evaluate the neighbouring points. A* does not generate the whole path at once, but iteratively checks the neighbours, and then adds the best-fitting neighbours to the path. The multi-fitness function $F$ that we use for the evaluation is constructed in such a way that it can only evaluate paths but not points (as in classical A* and as the cost function in Eq. (4) does).

To use the multi-fitness function $F$ for improved A*, it should restore the path already explored. This means that by evaluating the path to the neighbour point, improved A* collects the points, that the algorithm has already chosen, in one path sequence. The last point of this sequence will be the neighbour $n_i$, that is currently being evaluated. The multi-fitness function $F$ is then applied to evaluate the reconstructed path. The fitness value obtained is assigned to the point $n_i$. After that, the improved A* algorithm continues to work as its classical version (as shown in Fig.16).

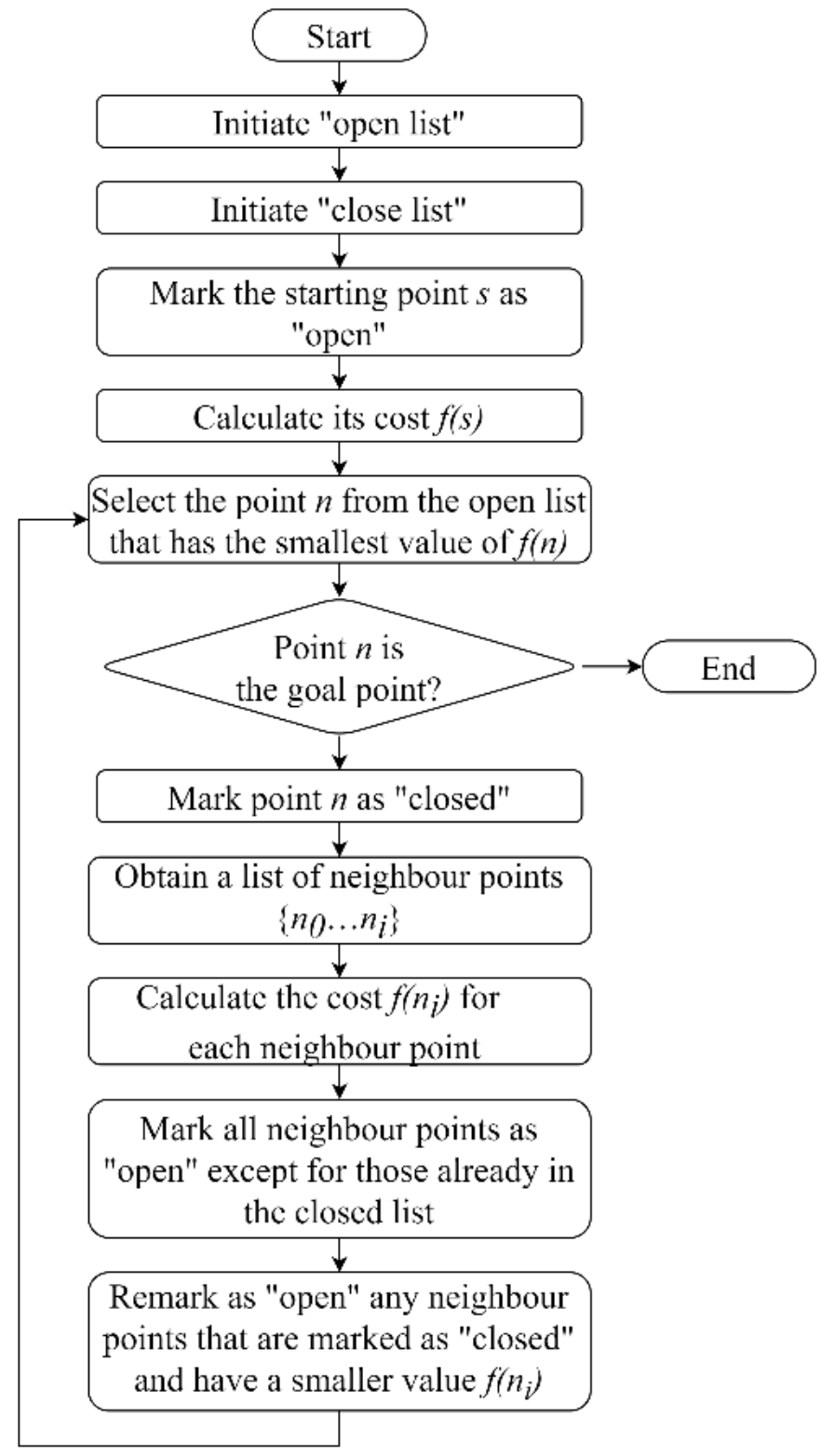

**Fig. 16.** Classical A* algorithm

### 4.7 Implemented Combination of Path Planning and Decision-Making Algorithms

In our experiments we combined two path planning algorithms, GA and A*, with the decision-making algorithm AHP. As result, we got two improved path planning algorithms: GA with the decision-making algorithm AHP (further denoted as GA–AHP) and A* with the decision-making algorithm AHP (further denoted as A*–AHP). In the rest of this paper, we denote these combinations just “improved algorithms”. Fig.17 summarizes the information provided above in one graph illustrating the combination of the selected algorithms.

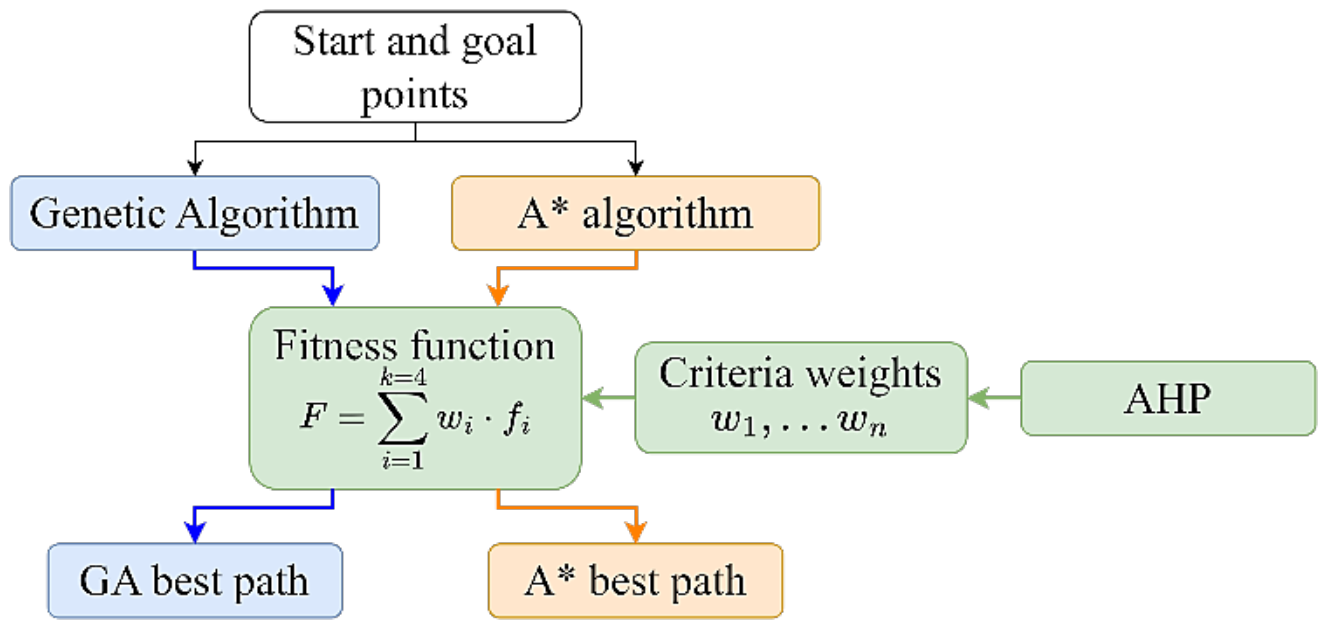

**Fig. 17.** Implemented combination of path planning and decision-making algorithms

Both path planning algorithms GA and A* receive the same pair of start and goal points and start to generate the paths. During path generation, each algorithm uses the multi-fitness function from Eq. (3) for path evaluation. GA uses this fitness function to evaluate each path in the population for each generation loop (see Fig.15). A* uses the fitness function each time, when it needs to calculate costs to the next neighbour (see Fig.16).

After the path planning algorithms finish their work, there is one path selected by each algorithm. For GA it is one of the paths from the population with the best (smallest) fitness value. A*, due to its structure, always provides only one path as output. This path is considered as the best one for A*.

The important aspect is that both algorithms use the same multi-fitness function i.e. Eq. (3) to evaluate the paths. This ensures fair comparison of the results of both approaches.

## 5 Experiments and Results

On the one hand, a multi-sections robot can be easily simulated, i.e. its environment can be constructed similar to the 2-sections model, whereby the closer to the “surface” the higher is the density of the environment i.e. the more points belong to different local environments but have the same spatial coordinates. These points can exactly be reached by different positions of the robot’s arm, i.e. without any accuracy deviation. This feature is especially interesting in the context of decision-making, when the robot should decide in which position it should move to reach the point. On the other hand, the multi-sections robot model does not take into account important physical features of a multi-sections robot. Each additional section in the physical prototype increases not only flexibility of the robot but also its weight significantly. For a real robot it will be harder to control the motions and ensure their accuracy. The path generation process will require much more time and energy. Each additional section of a robot leads to an expansion of the entire set-up, i.e. an increased number of motors, tendons’ length, etc.

Thus, the multi-sections robot is much easier to implement as a virtual model, than as a physical prototype. How simplifying the physical robot's features influences the results of the investigations and if the conclusions will be still applicable to a physical robot has experimentally to be investigated in the future.

For our experiments we used two different virtual environments (Environment 1 and Environment 2) as well as two configurations for the global environments (2-sections robot and 4-sections robot). To evaluate the paths generated by improved GA and improved A* depending on the multiple criteria applied, we analysed the output of both algorithms to discover which one manages to get better results. In this section we discuss the impact of different combinations of criteria weights on the path selection and performance of the improved path planning algorithms.

### 5.1 Impact of Different Combinations of Criteria Weights on the Path Selection

Usually, there are many motion trajectories between start and goal points in the global environment of the robot. The path planning algorithms, influenced by the criteria weights, generate the paths for each section of the robot from its start point to its goal point (i.e. between start and goal points in the local environment of the robot's section) and calculate the fitness value of the path using the multi-fitness function $F$. $F$ consists of 4 single fitness functions for distance, motor damage, mechanical damage and accuracy, and the weights of the applied criteria (see Eq. (3)). The multi-fitness function influences each path during its generation. Different criteria weights affect the fitness value of the path and as a consequence affect the output of the path planning algorithms. That means that, even for the same start and goal points, the algorithms generate different paths, depending on the criteria weights. In this sub-section, we describe our experiments aiming to show how the output of GA and A* path planning algorithms depends on the applied combination of the criteria weights.

For the 4 criteria – distance, motor damage, mechanical damage and accuracy – we defined and applied 15 different combinations of criteria weights as shown in Table 3. All combinations are numbered. Each line in Table 3 represents a single combination and each cell is indicating the weight for the corresponding criterion. The green cells indicate the prioritised criteria, i.e. the criteria with a significantly higher weight value. The weights were calculated according to their relative importance value (see section 4.1). Table 3 represents all possible combinations of 4 criteria excluding the case where no criteria are prioritised. Whether no or all criteria are prioritized, they will have an equal relative importance level i.e. the case in which no criteria are prioritised is equivalent to combination number 15.

**Table 3.** Combination of criteria weights (prioritized criteria are highlighted in light green)

| Index of combina-tion | Distance ($w_{dist}$) | Motors damage ($w_{mot}$) | Mechanical damage ($w_{mech}$) | Accuracy ($w_{acc}$) |
|---|---|---|---|---|
| ***Single criterion*** | | | | |
| 1 | 0.75 | 0.083 | 0.083 | 0.083 |
| 2 | 0.083 | 0.75 | 0.083 | 0.083 |
| 3 | 0.083 | 0.083 | 0.75 | 0.083 |
| 4 | 0.083 | 0.083 | 0.083 | 0.75 |
| ***Two criteria*** | | | | |
| 5 | 0.45 | 0.45 | 0.05 | 0.05 |
| 6 | 0.45 | 0.05 | 0.45 | 0.05 |
| 7 | 0.45 | 0.05 | 0.05 | 0.45 |
| 8 | 0.05 | 0.45 | 0.45 | 0.05 |
| 9 | 0.05 | 0.45 | 0.05 | 0.45 |
| 10 | 0.05 | 0.05 | 0.45 | 0.45 |
| ***Three criteria*** | | | | |
| 11 | 0.321 | 0.321 | 0.321 | 0.036 |
| 12 | 0.321 | 0.321 | 0.036 | 0.321 |
| 13 | 0.321 | 0.036 | 0.321 | 0.321 |
| 14 | 0.036 | 0.321 | 0.321 | 0.321 |
| ***Four criteria*** | | | | |
| 15 | 1 | 1 | 1 | 1 |

We investigated the impact of different combinations of criteria weights on the path selection for the 2- and 4-sections robot in different environments. For the 4-sections robot we selected randomly start and goal points from the robot's global Environment 1. The start point is represented with its index coordinates is {50, 47, 30, 30} and the goal point is {3, 14, 60, 29}. We selected the indices of two lower sections tips of the robot as the start and goal point for the 2-sections robot model, i.e. the start point is {50, 47}, and the goal point is {3, 14}. The same indices for start and goal points were used for the 2- and 4-sections robot for Environment 2. Fig.18 shows the start and goal points for the 2-sections robot for Environment 1 and Environment 2. Green points represent the local environment of the lower section. Grey points represent the local environment of the upper section for the start position. Red points represent the local environment of the upper section for the goal position. Fig.18 shows that the points with the same indices lead to the different spatial points in different global environment positions and, therefore, to different positions of the robot sections.

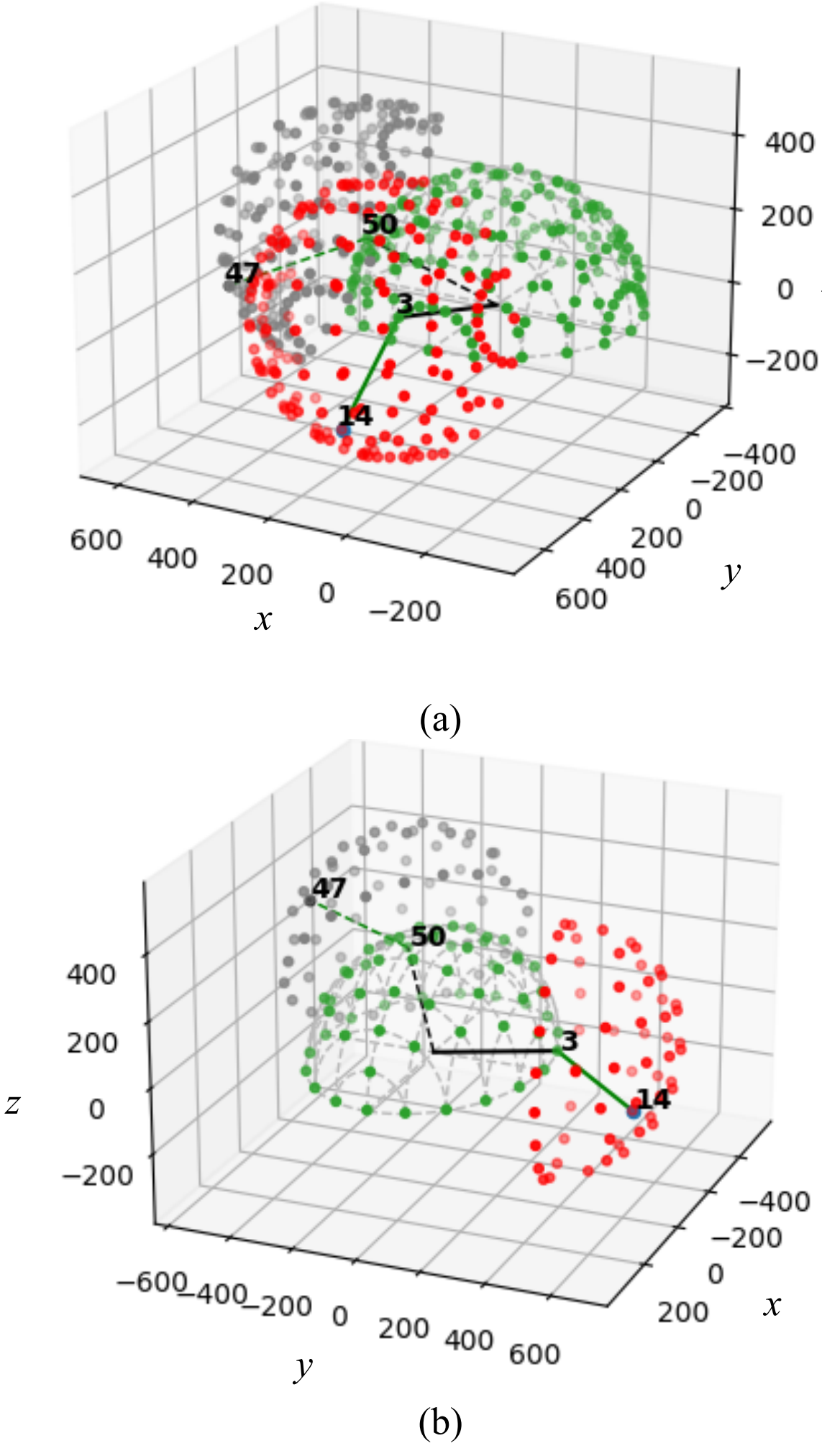

**Fig. 18.** Start and goal points and states of the 2-sections robot in our experiments in: (a) Environment 1, (b) Environment 2

The improved GA and improved A* algorithms plan the path for each robot's section using a fitness function and a certain combination of criteria weights, so that the robot can move from the starting position to the goal one. Algorithm 1 describes the steps of the path generation and fitness calculation. The generated path will be the best possible one for the applied combination of the criteria weights. The generation of such a "best" path is now an inherent/intrinsic feature of the combination of path planning and decision-making algorithms implemented here.

**Algorithm 1**: Path generation and calculation of the fitness value.

***input:*** *start_point* and *goal_point* as indice coordinates
***for*** *each_criteria_combination* // 15 combinations
  ***for*** *each_robot's_section* // 2 or 4 sections
    ***for*** *each_Path_Planning_algorithm* // *GA* and *A**
      generate the *path* // considering criteria weights
      compute *fitness_value* // using multi-fitness function *F*
    ***end for***
  accumulate *fitness_values*
  ***end for***
***output:*** *planned_path_GA_all_sections* and its *fitness_value;*
  *planned_path_A*_all_sections* and its *fitness_value.*
***end for***

For the improved GA algorithm, we chose a population size of 50 (number of chromosomes) and a generation number of 3 as values to conduct the experiment[4]. The maximum length of a chromosome was defined as 13. The generated paths were evaluated with the multi fitness function *F*.

After the execution of Algorithm 1, using a virtual environment, we obtained 2·30=60 paths for the 2-sections robot, one for each section, and 30 fitness values, i.e. one per complete path. For the 4-sections robot, Algorithm 1 generated 4·30=120 paths and 30 fitness values. Table 4 shows, as an example, all the fitness values calculated for all executions of the path planning algorithms depending on the applied combination of criteria weights numbered from 1 to 15 corresponding to Table 3. Additionally, the cells in Table 4 marked with the same colours show cases with the same paths. To increase the readability of the paper the generated paths are shown in the Appendix, in Fig. A1 and Fig. A2.

The smallest i.e. the best fitness value for each criteria combination is marked bold in Table 4. They show which of the two path planning algorithms (within one Environment) generated the better path. The identical fitness values are underlined. The fitness value of identical paths is identical.

For the Environment 1 both algorithms provide a diversity of paths. For 9 out of 15 combinations A* generated better paths than GA. Paths for the combinations 7 and 12 are equal. For the 4-sections robot no equal paths were generated, A* generated paths with better fitness values for 12 out of 15 criteria combinations.

For the Environment 2, GA generated more diverse paths for the different criteria combinations, while A* always generates the same paths for the given start and goal points. For the 2-sections robot the A* shows better fitness values for all 15 criteria combinations compared to GA. Same results are obtained for the 4-sections robot as well.

[4] We ran several experiments to determine a reasonable population size and number of generations. We omit here the detailed explanation of these experiments for better readability. The numbers reported provided the best result in terms of path quality according to the fitness value and time taken.

**Table 4.** Fitness values for the generated paths

| Robot | Start and goal points | Combi-na-tion of cri-teria weights | Fitness values for the generated paths | | | |
|---|---|---|---|---|---|---|
| | | | Local Environment 1 $N$=165 points | | Local Environment 2 $N$=61 points | |
| | | | *GA* | *A** | *GA* | *A** |
| 2 - sections | Start point: {50, 47} Goal point: {3,14} | 1 | **1559.50** | 1583.53 | 2139.27 | **1824.71** |
| | | 2 | 4793.19 | **4783.54** | 4751.13 | **4744.52** |
| | | 3 | 2251.92 | **2117.54** | 1593.53 | **1447.17** |
| | | 4 | **781.85** | 821.89 | 771.37 | **739.18** |
| | | 5 | **3343.49** | 3357.36 | 3508.49 | **3501.87** |
| | | 6 | 1929.73 | **1777.00** | 1576.49 | **1569.48** |
| | | 7 | 951.76 | 951.76 | 1202.79 | **1096.27** |
| | | 8 | 3792.39 | **3677.98** | 3351.34 | **3275.47** |
| | | 9 | **2876.60** | 2901.52 | 2886.72 | **2850.89** |
| | | 10 | 1352.60 | **1271.98** | 875.59 | **872.49** |
| | | 11 | 3023.79 | **2983.87** | 2980.19 | **2835.56** |
| | | 12 | 2395.65 | 2395.65 | 2771.59 | **2498.41** |
| | | 13 | 1430.07 | **1269.60** | 1269.83 | **1121.57** |
| | | 14 | 2681.25 | **2624.04** | 2368.42 | **2338.98** |
| | | 15 | 9419.91 | **9295.55** | 9302.35 | **8833.51** |
| 4 - sections | Start point: {50, 47, 30, 30} Goal point: {3, 14, 60, 29} | 1 | **3071.45** | 3210.19 | 2633.28 | **2318.71** |
| | | 2 | 5290.91 | **5171.39** | 4865.62 | **4804.30** |
| | | 3 | 4629.42 | **4243.59** | 1603.73 | **1508.95** |
| | | 4 | **1338.61** | 1342.83 | 863.19 | **794.96** |
| | | 5 | **4253.76** | 4341.49 | 3901.38 | **3800.68** |
| | | 6 | 4319.13 | **3803.52** | 2066.06 | **1869.49** |
| | | 7 | 2038.92 | **1928.29** | 1393.39 | **1392.68** |
| | | 8 | 5449.86 | **4961.59** | 3379.61 | **3315.07** |
| | | 9 | 3296.82 | **3225.73** | 2900.21 | **2886.89** |
| | | 10 | 3248.75 | **2547.99** | 982.42 | **909.69** |
| | | 11 | 4814.88 | **4434.88** | 3437.46 | **3051.27** |
| | | 12 | 3193.59 | **3098.54** | 2907.75 | **2711.56** |
| | | 13 | 3258.58 | **2715.19** | 1626.39 | **1335.57** |
| | | 14 | 4028.81 | **3540.31** | 2433.04 | **2367.45** |
| | | 15 | 14481.73 | **13815.83** | 10122.37 | **9505.52** |

The fitness values of the paths generated for the Environment 2 are in average 23.3% smaller for the 2-section robot and 20.6% smaller for the 4-sections robot, than for the paths generated for the Environment 1. This happens due to the different cardinality of the environments. Environment 2 has almost 3 times less points than Environment 1. This affects the length of the generated paths and also influences their fitness values.

Fig. 19 and Fig. 20 show a comparison of the fitness values for the generated paths for 100 executions of Algorithm 1 with randomly selected goal points. The initial position of the robot at the begging of the experiment is straight up, the start point of the very first path is equal to $N/2$. A start point of each next generated path is equal to the goal point of the previous generated path, so the robot does not come back to the initial position after each generated path, but continue to move from the point where it stopped. The results are represented in a histogram, where the black bars represent the number of cases in which A* provided paths with smaller fitness values than GA, the blue bars indicate cases where both algorithms generated paths with the same fitness value, and the grey bars show cases where GA outperformed A*. Each bar contains a value, and the sum of the three values in one column is always equal to 100 for each criteria combination.

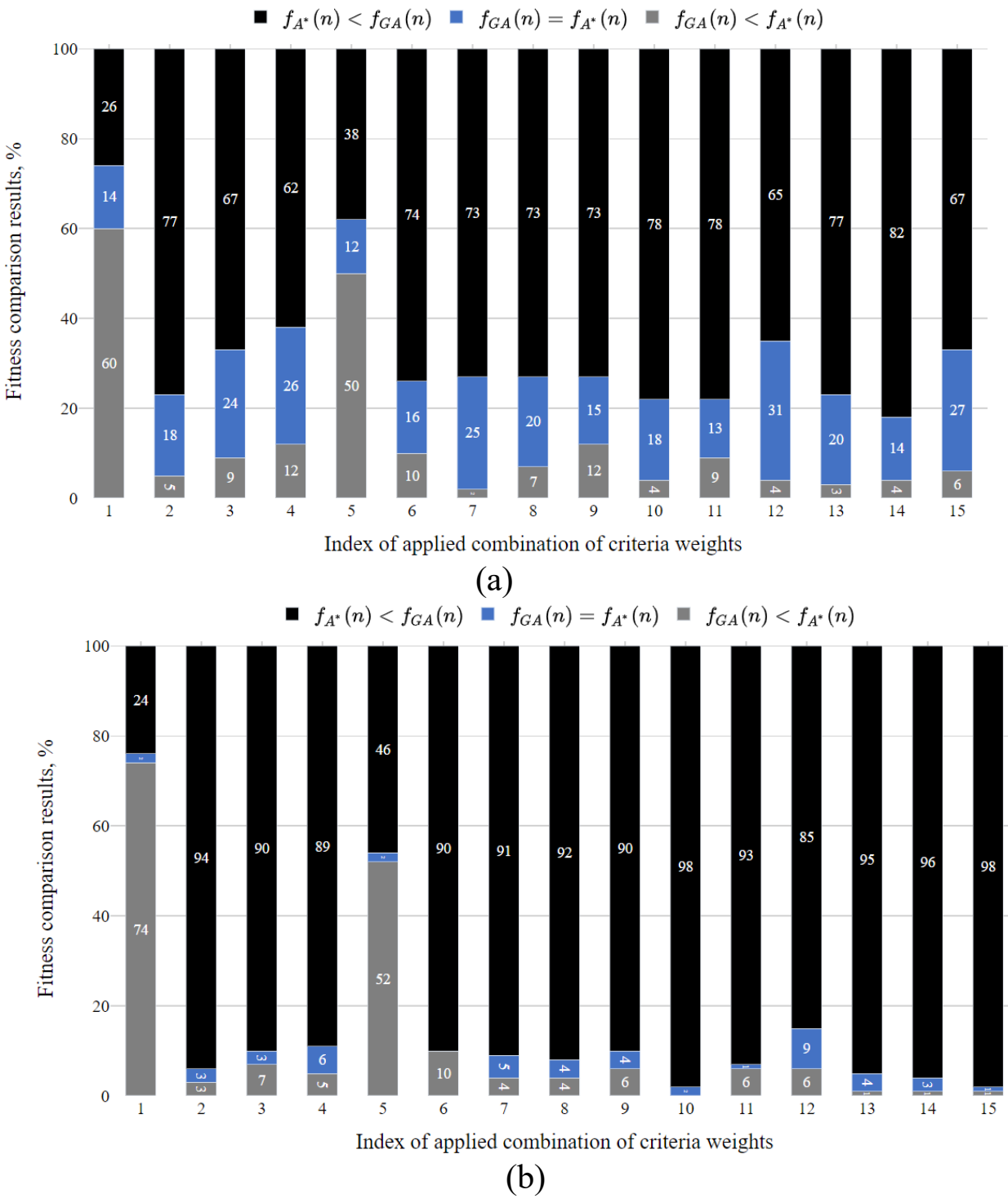

**Fig. 19.** Relation of solutions with better fitness values for the Environment 1: (a) 2-sections robot, (b) 4-sections robot. GA better than A* (grey), both yielding the same results (blue), A* better than GA (black).

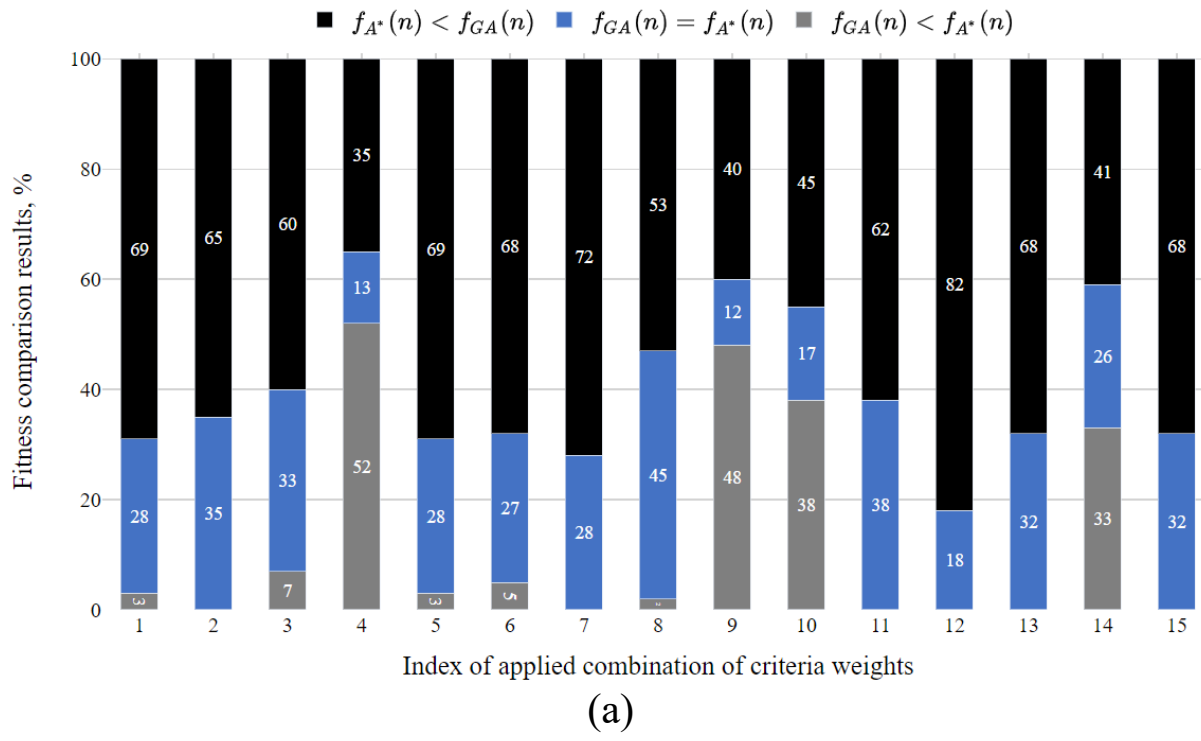

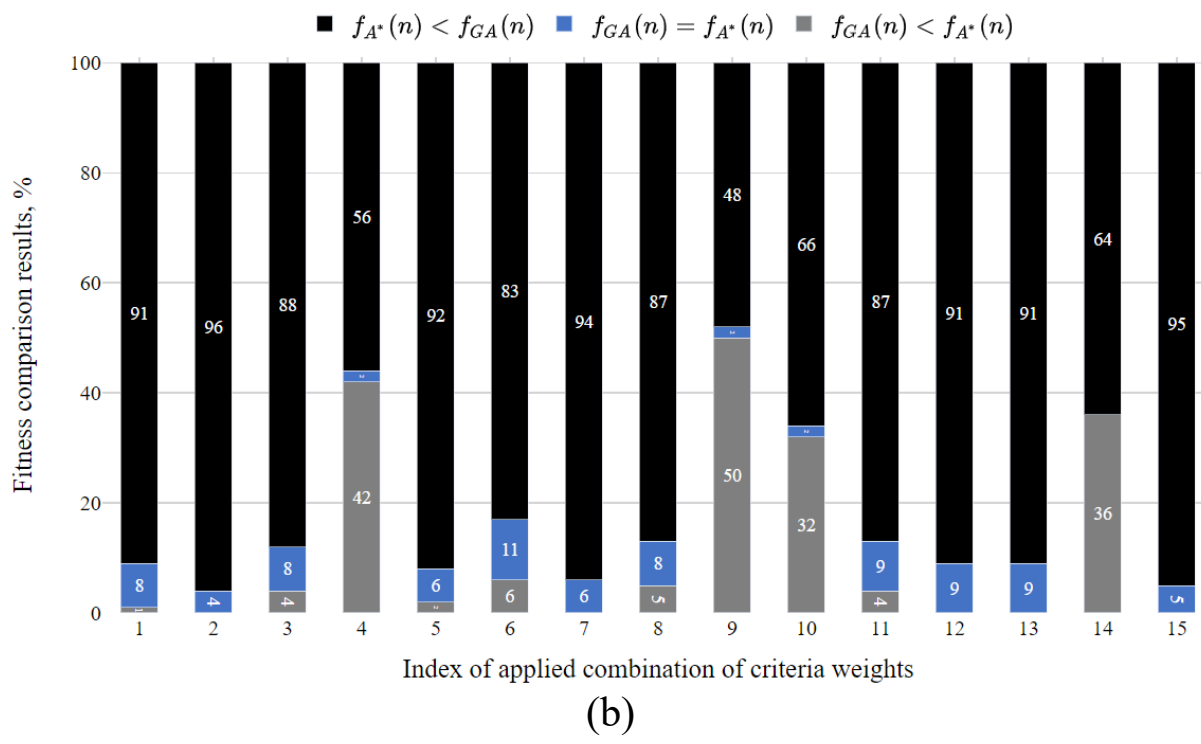

(b)

**Fig. 20.** Relation of solutions with better fitness values for the Environment 2: (a) 2-sections robot, (b) 4-sections robot.GA better than A* (grey), both yielding the same results (blue), A* better than GA (black).

For both environments, the results of the algorithms depend on the number of the robot's sections. In Environment 1, for the 2-and 4-section robot, A* generated more paths with better fitness values than GA for 13 out of 15 criteria combinations. In Environment 2, for both the 2- and 4-section robots, the number of same results is increased (blue columns). GA generated more paths with better fitness values for criteria combinations 4, 9, 10 and 14. Each of these four combinations includes prioritized "accuracy" criteria.

The results indicate that both algorithms possess distinct advantages and disadvantages. Despite the fact that the A* algorithm generates a greater number of paths with better fitness values, it does not provide a diversity of paths. A* tends to generate the same path for the same start and goal point, which leads to increased mechanical damage of the particular robot segments in long-term use of the robot and decrease the time to maintenance operations of the continuum robot. Unlike A*, GA generates more diverse paths, thus increasing the robot's resilience.

### 5.2 Impact of Different Combinations of Criteria Weights on the execution time of the Path Planning Algorithms

Table 5 shows the processing time of the algorithms for different virtual Environments and section numbers. Each cell in Table 5 contains an average value of 100 runs for the particular path planning algorithm and the particular criteria combination. The fastest processing time is shown by A* for the 2-sections robot for the Environment 2 (marked green). The slowest processing time is showed by GA for 4- sections robot for the Environment 2 (marked red). Please note, that the processing time of GA almost does not depend on the environment and as consequence on the number of points in the environment (cardinality), while the time performance of A* for Environment 1 (with 165 points) is almost 3 time higher than for Environment 2 (with 61 points). In the context of a real continuum robot characterised by an infinite number of sections and reachable points, we assume that GA would outperform A* algorithm due to its reduced dependence on the environment's cardinality in terms of execution time.

**Table 5.** Averaged processing time for the Algorithm 1

| Environment | Robot | Algorithm | Combination of criteria weights | | | | | | | | | | | | | | | Average | Median |
|---|---|---|---|---|---|---|---|---|---|---|---|---|---|---|---|---|---|---|---|
| | | | *1* | *2* | *3* | *4* | *5* | *6* | *7* | *8* | *9* | *10* | *11* | *12* | *13* | *14* | *15* | | |
| 1 | 2-sections | GA | 0.31 | 0.31 | 0.29 | 0.29 | 0.29 | 0.29 | 0.29 | 0.31 | 0.30 | 0.30 | 0.30 | 0.29 | 0.32 | 0.32 | 0.33 | 0.30 | 0.30 |
| | | A* | 0.53 | 0.36 | 0.50 | 0.10 | 0.46 | 0.50 | 0.17 | 0.51 | 0.10 | 0.25 | 0.52 | 0.17 | 0.36 | 0.26 | 0.33 | 0.34 | 0.36 |
| | 4-sections | GA | 0.60 | 0.62 | 0.63 | 0.60 | 0.61 | 0.62 | 0.61 | 0.64 | 0.62 | 0.64 | 0.63 | 0.62 | 0.63 | 0.63 | 0.63 | 0.62 | 0.62 |
| | | A* | 1.02 | 0.69 | 1.00 | 0.23 | 0.95 | 1.10 | 0.37 | 1.07 | 0.23 | 0.49 | 1.09 | 0.39 | 0.71 | 0.48 | 0.68 | 0.70 | 0.69 |
| 2 | 2-sections | GA | 0.40 | 0.40 | 0.40 | 0.40 | 0.41 | 0.40 | 0.40 | 0.42 | 0.40 | 0.38 | 0.39 | 0.42 | 0.39 | 0.40 | 0.41 | 0.40 | 0.40 |
| | | A* | 0.19 | 0.14 | 0.18 | 0.04 | 0.21 | 0.20 | 0.09 | 0.18 | 0.04 | 0.06 | 0.19 | 0.10 | 0.13 | 0.05 | 0.13 | 0.13 | 0.13 |
| | 4-sections | GA | 0.79 | 0.80 | 0.80 | 0.74 | 0.81 | 0.74 | 0.77 | 0.79 | 0.75 | 0.76 | 0.78 | 0.78 | 0.76 | 0.76 | 0.78 | 0.77 | 0.78 |
| | | A* | 0.40 | 0.26 | 0.29 | 0.07 | 0.43 | 0.36 | 0.16 | 0.27 | 0.07 | 0.10 | 0.38 | 0.17 | 0.22 | 0.10 | 0.24 | 0.23 | 0.24 |

# 6 Conclusion

Recently, resilience became a very important feature, due to the growing demand for autonomous systems. Such systems should be self-aware and have the ability to adapt their behaviour [2]. Path planning is considered as an essential task for autonomous systems and should be generated according to the system's and environment' s state. We used our self-made continuous robot prototype to research the impact of different criteria representing the robots current state influence the results of path planning algorithms.

This paper discusses an experiment with improved versions of GA and A* algorithm for two different robot configurations: 2- and 4-sections robots. We took classical versions of both algorithms and modified them by adding the decision-making algorithm AHP to allow for more criteria than the path length to be considered when assessing the quality of the path calculated. The modifications are aiming to increase the resilience properties of the system. The improved GA and A* algorithms plan the path using a fitness function and a certain combination of criteria weights, such as distance, robot's tip accuracy, motor damage, and mechanical damage. The generated path will be the best possible one for the applied combination of the criteria weights. The generation of such a "best" path is an inherent/intrinsic feature of the combination of path planning and decision-making algorithms implemented here. However, the execution time of the investigated path planning algorithms depends significantly not only on the applied combination of the criteria but also on the features of the robot's environment, i.e. on the number of the points in the environment and on whether all points are multi-path or not. This dependency has to be further investigated. A high number of points in the robot's environment can be a penalty for the performance of the algorithms in a real-word environment, especially for A* algorithm.

In order to improve resilience a high diversity of different paths selected is beneficial. GA generated more diverse paths for the different criteria combinations, in both environments, while A* tend to generate the same paths for the given start and goal points. The performance time of GA does not depend on the cardinality of the environment, in contrast to A*.

## References

1. Shamilyan O, Kabin I, Dyka Z, Langendoerfer P (2022) Distributed Artificial Intelligence as a Means to Achieve Self-X-Functions for Increasing Resilience: the First Steps. In: 2022 11th Mediterranean Conference on Embedded Computing (MECO). pp 1–6
2. Shamilyan O, Kabin I, Dyka Z, et al (2023) Intelligence and Motion Models of Continuum Robots: An Overview. IEEE Access 11:60988–61003. https://doi.org/10.1109/ACCESS.2023.3286300
3. Shamilyan O, Kabin I, Dyka Z, et al (2021) Octopuses: biological facts and technical solutions. In: 2021 10th Mediterranean Conference on Embedded Computing (MECO). IEEE, Budva, Montenegro, pp 1–7
4. Shamilyan O, Kabin I, Dyka Z, Langendoerfer P (2024) Resilient Movement Planning for Continuum Robots. In: Multi-Objective Decision Making Workshop at ECAI 2024. Santiago de Compostela, Spain
5. Nguyen T-D, Burgner-Kahrs J (2015) A tendon-driven continuum robot with extensible sections. In: 2015 IEEE/RSJ International Conference on Intelligent Robots and Systems (IROS). IEEE, pp 2130–2135
6. Renda F, Boyer F, Dias J, Seneviratne L (2018) Discrete Cosserat Approach for Multisection Soft Manipulator Dynamics. IEEE transactions on robotics : a publication of the IEEE Robotics and Automation Society 34:1518–1533. https://doi.org/10.1109/TRO.2018.2868815
7. Neumann M, Burgner-Kahrs J (2016) Considerations for follow-the-leader motion of extensible tendon-driven continuum robots. In: 2016 IEEE International Conference on Robotics and Automation (ICRA 2016). IEEE, Piscataway, NJ, pp 917–923
8. Rucker DC, Jones BA, Webster RJ (2010) A Geometrically Exact Model for Externally Loaded Concentric-Tube Continuum Robots. IEEE transactions on robotics : a publication of the IEEE Robotics and Automation Society 26:769–780. https://doi.org/10.1109/TRO.2010.2062570
9. Rucker DC, Webster III RJ (2011) Statics and Dynamics of Continuum Robots With General Tendon Routing and External Loading. IEEE transactions on robotics : a publication of the IEEE Robotics and Automation Society 27:1033–1044. https://doi.org/10.1109/TRO.2011.2160469
10. Renda F, Giorelli M, Calisti M, et al (2014) Dynamic Model of a Multibending Soft Robot Arm Driven by Cables. IEEE transactions on robotics : a publication of the IEEE Robotics and Automation Society 30:1109–1122. https://doi.org/10.1109/TRO.2014.2325992
11. Cosserat F, Cosserat E (1909) Théorie des Corps déformables. Nature 81:67. https://doi.org/10.1038/081067a0

12. Chikhaoui MT, Lilge S, Kleinschmidt S, Burgner-Kahrs J (2019) Comparison of Modeling Approaches for a Tendon Actuated Continuum Robot With Three Extensible Segments. IEEE Robotics and Automation Letters 4:989–996. https://doi.org/10.1109/LRA.2019.2893610
13. Thuruthel TG, Falotico E, Renda F, Laschi C (2017) Learning dynamic models for open loop predictive control of soft robotic manipulators. Bioinspiration & biomimetics 12:066003. https://doi.org/10.1088/1748-3190/aa839f
14. Thuruthel TG, Falotico E, Renda F, et al (2019) Emergence of behavior through morphology: a case study on an octopus inspired manipulator. Bioinspiration & biomimetics 14:034001. https://doi.org/10.1088/1748-3190/ab1621
15. George Thuruthel T, Ansari Y, Falotico E, Laschi C (2018) Control Strategies for Soft Robotic Manipulators: A Survey. Soft robotics 5:149–163. https://doi.org/10.1089/soro.2017.0007
16. Tang Z, Ma H (2021) An overview of path planning algorithms. IOP Conf Ser: Earth Environ Sci 804:022024. https://doi.org/10.1088/1755-1315/804/2/022024
17. Karur K, Sharma N, Dharmatti C, Siegel JE (2021) A Survey of Path Planning Algorithms for Mobile Robots. Vehicles 3:448–468. https://doi.org/10.3390/vehicles3030027
18. Gade PK, Osuri M (2014) Evaluation of Multi Criteria Decision Making Methods for Potential Use in Application Security
19. Aruldoss M, Lakshmi TM, Venkatesan VP (2013) A Survey on Multi Criteria Decision Making Methods and Its Applications. American Journal of Information Systems 1:31–43. https://doi.org/10.12691/ajis-1-1-5
20. Velasquez M, Hester PT (2013) An Analysis of Multi-Criteria Decision Making Methods
21. Kim C, Kim Y, Yi H (2020) Fuzzy Analytic Hierarchy Process-Based Mobile Robot Path Planning. Electronics 9:290. https://doi.org/10.3390/electronics9020290
22. Kim C, Suh J, Han J-H (2020) Development of a Hybrid Path Planning Algorithm and a Bio-Inspired Control for an Omni-Wheel Mobile Robot. Sensors 20:4258. https://doi.org/10.3390/s20154258
23. Multi-Criteria Decision Method Applied to Path Planning for Mobile Robots | IEEE Conference Publication | IEEE Xplore. https://ieeexplore.ieee.org/document/9995976. Accessed 29 Nov 2024
24. Zagradjanin, Pamucar, Jovanovic (2019) Cloud-Based Multi-Robot Path Planning in Complex and Crowded Environment with Multi-Criteria Decision Making using Full Consistency Method. Symmetry 11:1241. https://doi.org/10.3390/sym11101241
25. Damos MA, Zhu J, Li W, et al (2021) A Novel Urban Tourism Path Planning Approach Based on a Multiobjective Genetic Algorithm. IJGI 10:530. https://doi.org/10.3390/ijgi10080530
26. Wang H (2022) Continuum Robot Path Planning Based on Improved Genetic Algorithm. In: 2022 2nd International Conference on Algorithms, High Performance Computing and Artificial Intelligence (AHPCAI). pp 23–29

27. García-Cascales MS, Lamata MT, Verdegay JL (2010) The TOPSIS Method and Its Application to Linguistic Variables. In: Greco S, Marques Pereira RA, Squillante M, et al (eds) Preferences and Decisions: Models and Applications. Springer, Berlin, Heidelberg, pp 383–395
28. Saaty TL (1980) The Analytic Hierarchy Process: Planning, Priority Setting, Resource Allocation. McGraw-Hill International Book Company
29. AHP calculator - AHP-OS. https://bpmsg.com/ahp/ahp-calc.php. Accessed 29 Feb 2024
30. Analytic Hierarchy Process (AHP) Tool. https://comcastsamples.github.io/ahp-tool/. Accessed 29 Feb 2024
31. Tutorials - AHP Calculation Methods. https://www.spicelogic.com/docs/ahpsoftware/intro/ahp-calculation-methods-396. Accessed 29 Feb 2024
32. Sharma P (2020) Understanding Distance Metrics Used in Machine Learning. In: Analytics Vidhya. https://www.analyticsvidhya.com/blog/2020/02/4-types-of-distance-metrics-in-machine-learning/. Accessed 7 Mar 2024
33. Adaptation in Natural and Artificial Systems. In: MIT Press. https://mitpress.mit.edu/9780262581110/adaptation-in-natural-and-artificial-systems/. Accessed 1 Aug 2023
34. Bhoskar MsT, Kulkarni MrOK, Kulkarni MrNK, et al (2015) Genetic Algorithm and its Applications to Mechanical Engineering: A Review. Materials Today: Proceedings 2:2624–2630. https://doi.org/10.1016/j.matpr.2015.07.219
35. Griffiths IJ, Mehdi QH, Wang T, Gough NE (1997) A Genetic Algorithm for Path Planning. IFAC Proceedings Volumes 30:485–490. https://doi.org/10.1016/S1474-6670(17)43312-X
36. Latin Hypercube Sampling with a Genetic Algorithm — geneticLHS. https://bertcarnell.github.io/lhs/reference/geneticLHS.html. Accessed 5 Dec 2024
37. Katoch S, Chauhan SS, Kumar V (2021) A review on genetic algorithm: past, present, and future. Multimed Tools Appl 80:8091–8126. https://doi.org/10.1007/s11042-020-10139-6
38. Luca GD (2020) Roulette Selection in Genetic Algorithms | Baeldung on Computer Science. https://www.baeldung.com/cs/genetic-algorithms-roulette-selection. Accessed 27 Feb 2024
39. Genetic Algorithms - Mutation. https://www.tutorialspoint.com/genetic_algorithms/genetic_algorithms_mutation.htm. Accessed 27 Feb 2024
40. Hart PE, Nilsson NJ, Raphael B (1968) A Formal Basis for the Heuristic Determination of Minimum Cost Paths. IEEE Transactions on Systems Science and Cybernetics 4:100–107. https://doi.org/10.1109/TSSC.1968.300136
41. Barnouti NH, Al-Dabbagh SSM, Sahib Naser MA (2016) Pathfinding in Strategy Games and Maze Solving Using A* Search Algorithm. JCC 04:15–25. https://doi.org/10.4236/jcc.2016.411002
42. Guruji AK, Agarwal H, Parsediya DK (2016) Time-efficient A* Algorithm for Robot Path Planning. Procedia Technology 23:144–149. https://doi.org/10.1016/j.protcy.2016.03.010

43. Zuo L, Guo Q, Xu X, Fu H (2015) A hierarchical path planning approach based on A∗ and least-squares policy iteration for mobile robots. Neurocomputing 170:257–266. https://doi.org/10.1016/j.neucom.2014.09.092
44. Jianqin L, Xiao G (2022) Research on improved A-star algorithm for global path planning of unmanned logistics vehicles. In: 2022 14th International Conference on Intelligent Human-Machine Systems and Cybernetics (IHMSC). pp 44–47

# Appendix

| Combination | GA path | | A* path | |
|---|---|---|---|---|
| | section_1 | section_2 | section_1 | section_2 |
| 1 | 50, 72, 79, 78, 75, 71, 67, 63, 52, 53, 54, 55, 56 | 47, 29, 25, 14 | 50, 72, 79, 80, 81, 64, 43, 26, 12, 1, 2, 3 | 47, 29, 25, 14 |
| 2 | 50, 72, 79, 80, 81, 64, 43, 26, 12, 1, 2, 3 | 47, 29, 25, 14 | 50, 72, 79, 80, 81, 82, 57, 36, 19, 6, 5, 4, 3 | 47, 29, 25, 14 |
| 3 | 50, 72, 79, 80, 81, 64, 43, 26, 12, 1, 2, 3 | 47, 29, 25, 14 | 50, 72, 79, 80, 81, 82, 57, 36, 19, 6, 5, 4, 3 | 47, 29, 25, 14 |
| 4 | 50, 72, 79, 80, 81, 64, 43, 26, 12, 1, 2, 3 | 47, 29, 25, 14 | 50, 46, 42, 31, 32, 33, 34, 35, 36, 19, 6, 5, 4, 3 | 47, 29, 25, 14 |
| 5 | 50, 72, 79, 78, 75, 71, 67, 63, 52, 53, 54, 55, 56 | 47, 29, 25, 14 | 50, 72, 79, 80, 81, 64, 43, 26, 12, 1, 2, 3 | 47, 29, 25, 14 |
| 6 | 50, 72, 79, 80, 81, 82, 107, 128, 129, 130, 131, 132, 133 | 47, 29, 25, 14 | 50, 72, 79, 80, 81, 82, 57, 36, 19, 6, 5, 4, 3 | 47, 29, 25, 14 |
| 7 | 50, 72, 79, 80, 81, 64, 43, 26, 12, 1, 2, 3 | 47, 29, 25, 14 | 50, 72, 79, 80, 81, 64, 43, 26, 12, 1, 2, 3 | 47, 29, 25, 14 |
| 8 | 50, 72, 79, 80, 81, 82, 83, 65, 44, 27, 13, 11, 10 | 47, 29, 25, 14 | 50, 72, 79, 80, 81, 82, 57, 36, 19, 6, 5, 4, 3 | 47, 29, 25, 14 |
| 9 | 50, 72, 79, 80, 81, 64, 43, 26, 12, 1, 2, 3 | 47, 29, 25, 14 | 50, 46, 42, 31, 32, 33, 34, 35, 36, 19, 6, 5, 4, 3 | 47, 29, 25, 14 |
| 10 | 50, 72, 79, 80, 81, 64, 43, 26, 12, 1, 2, 3 | 47, 29, 25, 14 | 50, 72, 79, 80, 81, 82, 57, 36, 19, 6, 5, 4, 3 | 47, 29, 25, 14 |
| 11 | 50, 72, 79, 80, 81, 64, 43, 26, 12, 1, 2, 3 | 47, 29, 25, 14 | 50, 72, 79, 80, 81, 82, 57, 36, 19, 6, 5, 4, 3 | 47, 29, 25, 14 |
| 12 | 50, 72, 79, 80, 81, 64, 43, 26, 12, 1, 2, 3 | 47, 29, 25, 14 | 50, 72, 79, 80, 81, 64, 43, 26, 12, 1, 2, 3 | 47, 29, 25, 14 |
| 13 | 50, 46, 42, 31, 32, 33, 34, 35, 36, 19, 6, 7, 8 | 47, 29, 25, 14 | 50, 72, 79, 80, 81, 82, 57, 36, 19, 6, 5, 4, 3 | 47, 29, 25, 14 |
| 14 | 50, 72, 79, 80, 81, 64, 43, 26, 12, 1, 2, 3 | 47, 29, 25, 14 | 50, 72, 79, 80, 81, 82, 57, 36, 19, 6, 5, 4, 3 | 47, 29, 25, 14 |
| 15 | 50, 72, 79, 80, 81, 64, 43, 26, 12, 1, 2, 3 | 47, 29, 25, 14 | 50, 72, 79, 80, 81, 82, 57, 36, 19, 6, 5, 4, 3 | 47, 29, 25, 14 |

(a)

| Combination | GA path | | | |
|---|---|---|---|---|
| | section_1 | section_2 | section_3 | section_4 |
| 1 | 50, 72, 79, 78, 75, 71, 67, 63, 52, 53, 54, 55, 56 | 47, 29, 25, 14 | 30, 28, 24, 23, 22, 21, 20, 19, 6, 7, 8, 9, 10 | 30, 28, 24, 23, 22, 21, 20, 19, 6, 7, 8, 9, 10 |
| 2 | 50, 72, 79, 80, 81, 64, 43, 26, 12, 1, 2, 3 | 47, 29, 25, 14 | 30, 48, 69, 84, 85, 86, 76, 74, 70, 66, 62, 61, 60 | 30, 28, 24, 23, 22, 21, 20, 19, 6, 5, 4, 3, 2 |
| 3 | 50, 72, 79, 80, 81, 64, 43, 26, 12, 1, 2, 3 | 47, 29, 25, 14 | 30, 48, 69, 84, 83, 82, 57, 58, 59, 60 | 30, 48, 69, 84, 83, 82, 107, 128, 145, 144, 143, 142, 141 |
| 4 | 50, 72, 79, 80, 81, 64, 43, 26, 12, 1, 2, 3 | 47, 29, 25, 14 | 30, 48, 69, 84, 85, 86, 76, 74, 70, 66, 62, 61, 60 | 30, 28, 24, 23, 22, 21, 20, 19, 36, 35, 34, 33, 32 |
| 5 | 50, 72, 79, 78, 75, 71, 67, 63, 52, 53, 54, 55, 56 | 47, 29, 25, 14 | 30, 28, 24, 23, 22, 21, 20, 19, 6, 7, 8, 9, 10 | 30, 28, 24, 23, 22, 21, 20, 19, 6, 5, 4, 3, 2 |
| 6 | 50, 72, 79, 80, 81, 64, 43, 26, 12, 1, 2, 3 | 47, 29, 25, 14 | 30, 48, 69, 84, 83, 82, 57, 36, 19, 20, 21, 22, 23 | 30, 28, 24, 23, 22, 21, 20, 19, 6, 5, 4, 3, 2 |
| 7 | 50, 72, 79, 80, 81, 64, 43, 26, 12, 1, 2, 3 | 47, 29, 25, 14 | 30, 28, 24, 23, 22, 21, 20, 19, 36, 57, 58, 59, 60 | 30, 28, 24, 23, 22, 21, 20, 19, 6, 5, 4, 3, 2 |
| 8 | 50, 72, 79, 80, 81, 82, 83, 100, 121, 138, 152, 163, 162 | 47, 29, 25, 14 | 30, 48, 69, 84, 83, 82, 81, 80, 79, 72, 50, 46, 42 | 30, 28, 24, 23, 22, 21, 20, 19, 36, 57, 82, 81, 80 |
| 9 | 50, 46, 42, 31, 32, 33, 34, 35, 36, 19, 18, 17, 16 | 47, 29, 25, 14 | 30, 28, 24, 23, 22, 21, 20, 19, 36, 57, 58, 59, 60 | 30, 28, 24, 23, 22, 21, 20, 19, 36, 35, 34, 33, 32 |
| 10 | 50, 72, 79, 80, 81, 64, 43, 26, 12, 1, 2, 3 | 47, 29, 25, 14 | 30, 48, 69, 84, 85, 86, 76, 74, 70, 66, 62, 61, 60 | 30, 28, 24, 23, 22, 21, 20, 19, 36, 57, 56, 55, 54 |
| 11 | 50, 72, 79, 80, 81, 64, 43, 26, 12, 1, 2, 3 | 47, 29, 25, 14 | 30, 28, 24, 23, 22, 21, 20, 19, 36, 57, 82, 83, 84 | 30, 28, 24, 23, 22, 21, 20, 19, 6, 5, 4, 3, 2 |
| 12 | 50, 72, 79, 80, 81, 64, 43, 26, 12, 1, 2, 3 | 47, 29, 25, 14 | 30, 48, 69, 84, 85, 86, 89, 93, 97, 101, 112, 111, 110 | 30, 28, 24, 23, 22, 21, 20, 19, 6, 5, 4, 3, 2 |
| 13 | 50, 46, 42, 31, 32, 33, 34, 35, 36, 19, 6, 5, 4 | 47, 29, 25, 14 | 30, 48, 69, 84, 85, 86, 76, 74, 70, 66, 62, 61, 60 | 30, 28, 24, 23, 22, 21, 20, 19, 6, 5, 4, 3, 2 |
| 14 | 50, 72, 79, 80, 81, 64, 43, 26, 12, 1, 2, 3 | 47, 29, 25, 14 | 30, 28, 24, 23, 22, 21, 20, 19, 36, 57, 58, 59, 60 | 30, 48, 69, 84, 83, 82, 81, 64, 43, 26, 12, 1, 2 |
| 15 | 50, 72, 79, 80, 81, 82, 57, 36, 19, 6, 5, 4, 3 | 47, 29, 25, 14 | 30, 48, 69, 84, 83, 82, 57, 58, 59, 60 | 30, 28, 24, 23, 22, 21, 20, 19, 6, 5, 4, 3, 2 |

| A* path | | | |
|---|---|---|---|
| section_1 | section_2 | section_3 | section_4 |
| 50, 72, 79, 80, 81, 64, 43, 26, 12, 1, 2, 3 | 47, 29, 25, 14 | 30, 48, 69, 84, 83, 82, 57, 58, 59, 60 | 30, 48, 69, 84, 83, 82, 81, 80, 68, 47, 29 |
| 50, 72, 79, 80, 81, 82, 57, 36, 19, 6, 5, 4, 3 | 47, 29, 25, 14 | 30, 48, 69, 84, 83, 82, 57, 58, 59, 60 | 30, 48, 69, 84, 83, 82, 81, 80, 68, 47, 29 |
| 50, 72, 79, 80, 81, 82, 57, 36, 19, 6, 5, 4, 3 | 47, 29, 25, 14 | 30, 48, 69, 84, 83, 82, 57, 58, 59, 60 | 30, 48, 69, 84, 83, 82, 81, 80, 68, 47, 29 |
| 50, 46, 42, 31, 32, 33, 34, 35, 36, 19, 6, 5, 4, 3 | 47, 29, 25, 14 | 30, 48, 69, 84, 85, 86, 76, 74, 70, 66, 62, 61, 60 | 30, 28, 24, 23, 22, 21, 20, 19, 18, 17, 16, 15, 14, 25, 29 |
| 50, 72, 79, 80, 81, 64, 43, 26, 12, 1, 2, 3 | 47, 29, 25, 14 | 30, 48, 69, 84, 83, 82, 57, 58, 59, 60 | 30, 48, 69, 84, 83, 82, 81, 80, 68, 47, 29 |
| 50, 72, 79, 80, 81, 82, 57, 36, 19, 6, 5, 4, 3 | 47, 29, 25, 14 | 30, 48, 69, 84, 83, 82, 57, 58, 59, 60 | 30, 48, 69, 84, 83, 82, 81, 80, 68, 47, 29 |
| 50, 72, 79, 80, 81, 64, 43, 26, 12, 1, 2, 3 | 47, 29, 25, 14 | 30, 48, 69, 84, 83, 82, 57, 58, 59, 60 | 30, 48, 69, 84, 83, 82, 81, 80, 68, 47, 29 |
| 50, 72, 79, 80, 81, 82, 57, 36, 19, 6, 5, 4, 3 | 47, 29, 25, 14 | 30, 48, 69, 84, 83, 82, 57, 58, 59, 60 | 30, 48, 69, 84, 83, 82, 81, 80, 68, 47, 29 |
| 50, 46, 42, 31, 32, 33, 34, 35, 36, 19, 6, 5, 4, 3 | 47, 29, 25, 14 | 30, 48, 69, 84, 85, 86, 76, 74, 70, 66, 62, 61, 60 | 30, 28, 24, 23, 22, 21, 20, 19, 18, 17, 16, 15, 14, 25, 29 |
| 50, 72, 79, 80, 81, 82, 57, 36, 19, 6, 5, 4, 3 | 47, 29, 25, 14 | 30, 48, 69, 84, 83, 82, 57, 58, 59, 60 | 30, 48, 69, 84, 83, 82, 81, 80, 68, 47, 29 |
| 50, 72, 79, 80, 81, 82, 57, 36, 19, 6, 5, 4, 3 | 47, 29, 25, 14 | 30, 48, 69, 84, 83, 82, 57, 58, 59, 60 | 30, 48, 69, 84, 83, 82, 81, 80, 68, 47, 29 |
| 50, 72, 79, 80, 81, 64, 43, 26, 12, 1, 2, 3 | 47, 29, 25, 14 | 30, 48, 69, 84, 83, 82, 57, 58, 59, 60 | 30, 48, 69, 84, 83, 82, 81, 80, 68, 47, 29 |
| 50, 72, 79, 80, 81, 82, 57, 36, 19, 6, 5, 4, 3 | 47, 29, 25, 14 | 30, 48, 69, 84, 83, 82, 57, 58, 59, 60 | 30, 48, 69, 84, 83, 82, 81, 80, 68, 47, 29 |
| 50, 72, 79, 80, 81, 82, 57, 36, 19, 6, 5, 4, 3 | 47, 29, 25, 14 | 30, 48, 69, 84, 83, 82, 57, 58, 59, 60 | 30, 48, 69, 84, 83, 82, 81, 80, 68, 47, 29 |
| 50, 72, 79, 80, 81, 82, 57, 36, 19, 6, 5, 4, 3 | 47, 29, 25, 14 | 30, 48, 69, 84, 83, 82, 57, 58, 59, 60 | 30, 48, 69, 84, 83, 82, 81, 80, 68, 47, 29 |

(b)

**Fig. A1.** Generated paths for the Algorithm 1in Environment 1 for (a) the 2-sections robot, (b) the 4-sections robot.
Each line shows the generated path for a particular criteria combination group ("Combination" from the Table 3). Same paths are highlighted with the same colours.

| Combination | GA path | | A* path | |
|---|---|---|---|---|
| | section_1 | section_2 | section_1 | section_2 |
| 1 | 50, 51, 43, 42, 32, 22, 21, 13, 7, 3 | 47, 39, 40, 30, 31, 32, 22, 14 | 50, 42, 32, 31, 21, 13, 7, 3 | 47, 48, 40, 30, 31, 32, 22, 14 |
| 2 | 50, 42, 32, 22, 14, 8, 7, 3 | 47, 39, 29, 19, 20, 21, 13, 14 | 50, 42, 32, 22, 14, 13, 7, 3 | 47, 39, 29, 19, 11, 12, 13, 14 |
| 3 | 50, 42, 32, 33, 23, 15, 14, 8, 7, 3 | 47, 53, 57, 58, 59, 55, 56, 50, 49, 48, 40, 41, 42 | 50, 42, 32, 22, 14, 8, 7, 3 | 47, 39, 29, 19, 11, 12, 13, 14 |
| 4 | 50, 42, 41, 40, 30, 31, 21, 13, 7, 3 | 47, 48, 49, 41, 31, 21, 13, 14 | 50, 42, 32, 22, 14, 8, 7, 3 | 47, 39, 29, 30, 31, 21, 13, 14 |
| 5 | 50, 42, 32, 22, 14, 13, 7, 3 | 47, 39, 29, 30, 20, 21, 13, 14 | 50, 42, 32, 31, 21, 13, 7, 3 | 47, 48, 40, 30, 31, 32, 22, 14 |
| 6 | 50, 49, 41, 31, 21, 13, 7, 3 | 47, 53, 57, 58, 59, 55, 49, 50, 42, 32, 31, 30, 40 | 50, 42, 32, 22, 14, 13, 7, 3 | 47, 39, 29, 19, 11, 12, 13, 14 |
| 7 | 50, 42, 32, 22, 21, 13, 7, 3 | 47, 48, 40, 41, 31, 30, 20, 21, 13, 14 | 50, 42, 32, 31, 21, 13, 7, 3 | 47, 48, 40, 30, 31, 32, 22, 14 |
| 8 | 50, 42, 32, 22, 23, 15, 14, 8, 7, 3 | 47, 39, 40, 30, 31, 21, 22, 14 | 50, 42, 32, 22, 14, 8, 7, 3 | 47, 39, 29, 19, 11, 12, 13, 14 |
| 9 | 50, 42, 32, 22, 14, 13, 7, 3 | 47, 48, 49, 50, 42, 41, 40, 30, 31, 21, 22, 14 | 50, 42, 32, 22, 14, 8, 7, 3 | 47, 39, 29, 30, 31, 21, 13, 14 |
| 10 | 50, 42, 32, 22, 14, 8, 7, 3 | 47, 39, 29, 30, 31, 32, 22, 14 | 50, 42, 32, 22, 14, 8, 7, 3 | 47, 39, 29, 30, 31, 21, 13, 14 |
| 11 | 50, 49, 41, 40, 30, 31, 21, 13, 7, 3 | 47, 39, 40, 30, 20, 21, 13, 14 | 50, 42, 32, 22, 14, 13, 7, 3 | 47, 39, 29, 19, 11, 12, 13, 14 |
| 12 | 50, 42, 43, 33, 32, 22, 23, 15, 14, 8, 7, 3 | 47, 48, 49, 41, 31, 21, 13, 14 | 50, 42, 32, 31, 21, 13, 7, 3 | 47, 48, 40, 30, 31, 32, 22, 14 |
| 13 | 50, 49, 41, 42, 32, 22, 14, 13, 7, 3 | 47, 39, 40, 30, 20, 21, 22, 14 | 50, 42, 32, 22, 14, 13, 7, 3 | 47, 39, 29, 19, 11, 12, 13, 14 |
| 14 | 50, 51, 43, 33, 23, 15, 14, 8, 7, 3 | 47, 48, 40, 30, 20, 21, 13, 14 | 50, 42, 32, 22, 14, 8, 7, 3 | 47, 39, 29, 30, 31, 21, 13, 14 |
| 15 | 50, 42, 43, 44, 34, 24, 23, 33, 32, 22, 41, 40, 39 | 47, 48, 49, 50, 42, 32, 22, 14 | 50, 42, 32, 22, 14, 13, 7, 3 | 47, 39, 29, 19, 11, 12, 13, 14 |

(a)

| Combination | GA path | | | |
|---|---|---|---|---|
| | section_1 | section_2 | section_3 | section_4 |
| 1 | 50, 42, 32, 31, 21, 22, 14, 8, 7, 3 | 47, 48, 40, 30, 31, 32, 22, 14 | 30, 40, 48, 54, 58, 60 | 30, 29 |
| 2 | 50, 49, 48, 40, 30, 20, 12, 6, 7, 3 | 47, 48, 40, 41, 31, 21, 22, 14 | 30, 31, 41, 40, 48, 54, 58, 60 | 30, 29 |
| 3 | 50, 56, 55, 54, 48, 40, 41, 15, 14, 22, 32, 42, 41 | 47, 39, 40, 30, 31, 21, 13, 14 | 30, 40, 48, 54, 58, 60 | 30, 29 |
| 4 | 50, 51, 43, 42, 32, 22, 14, 13, 7, 3 | 47, 39, 40, 30, 31, 32, 22, 14 | 30, 40, 48, 49, 55, 59, 58, 60 | 30, 29 |
| 5 | 50, 42, 41, 31, 21, 13, 7, 3 | 47, 39, 29, 30, 31, 21, 22, 14 | 30, 31, 32, 33, 34, 44, 43, 42, 41, 49, 55, 38, 46 | 30, 29 |
| 6 | 50, 42, 32, 22, 14, 8, 7, 3 | 47, 39, 40, 41, 31, 30, 20, 21, 13, 14 | 30, 40, 48, 54, 58, 60 | 30, 29 |
| 7 | 50, 42, 32, 31, 21, 13, 7, 3 | 47, 39, 40, 41, 31, 21, 13, 14 | 30, 40, 48, 54, 58, 60 | 30, 29 |
| 8 | 50, 51, 43, 44, 34, 24, 23, 15, 14, 8, 7, 3 | 47, 39, 40, 41, 31, 21, 22, 14 | 30, 40, 48, 54, 58, 60 | 30, 29 |
| 9 | 50, 42, 32, 22, 14, 8, 7, 3 | 47, 48, 40, 30, 31, 32, 22, 14 | 30, 29, 39, 47, 53, 54, 58, 60 | 30, 29 |
| 10 | 50, 42, 41, 31, 21, 13, 7, 3 | 47, 39, 40, 30, 20, 12, 13, 14 | 30, 40, 48, 47, 53, 57, 58, 60 | 30, 29 |
| 11 | 50, 49, 41, 31, 21, 13, 12, 6, 2, 3 | 47, 46, 38, 39, 29, 30, 40, 41, 31, 32, 22, 14 | 30, 40, 48, 54, 58, 60 | 30, 29 |
| 12 | 50, 42, 32, 22, 14, 8, 7, 3 | 47, 48, 49, 50, 51, 43, 33, 23, 22, 14 | 30, 40, 48, 54, 58, 60 | 30, 29 |
| 13 | 50, 42, 32, 33, 23, 22, 14, 8, 7, 3 | 47, 39, 29, 19, 20, 30, 31, 21, 22, 14 | 30, 40, 48, 54, 58, 60 | 30, 29 |
| 14 | 50, 42, 32, 22, 23, 15, 14, 8, 7, 3 | 47, 48, 49, 41, 42, 32, 22, 14 | 30, 31, 41, 49, 48, 54, 58, 60 | 30, 29 |
| 15 | 50, 42, 32, 22, 14, 13, 7, 3 | 47, 46, 38, 39, 40, 30, 31, 21, 13, 14 | 30, 40, 39, 47, 48, 54, 58, 60 | 30, 29 |

| A* path | | | |
|---|---|---|---|
| section_1 | section_2 | section_3 | section_4 |
| 50, 42, 32, 31, 21, 13, 7, 3 | 47, 48, 40, 30, 31, 32, 22, 14 | 30, 40, 48, 54, 58, 60 | 30, 29 |
| 50, 42, 32, 22, 14, 13, 7, 3 | 47, 39, 29, 19, 11, 12, 13, 14 | 30, 40, 48, 54, 58, 60 | 30, 29 |
| 50, 42, 32, 22, 14, 8, 7, 3 | 47, 39, 29, 19, 11, 12, 13, 14 | 30, 40, 48, 54, 58, 60 | 30, 29 |
| 50, 42, 32, 22, 14, 8, 7, 3 | 47, 39, 29, 30, 31, 21, 13, 14 | 30, 40, 48, 54, 58, 60 | 30, 29 |
| 50, 42, 32, 31, 21, 13, 7, 3 | 47, 48, 40, 30, 31, 32, 22, 14 | 30, 40, 48, 54, 58, 60 | 30, 29 |
| 50, 42, 32, 22, 14, 13, 7, 3 | 47, 39, 29, 19, 11, 12, 13, 14 | 30, 40, 48, 54, 58, 60 | 30, 29 |
| 50, 42, 32, 31, 21, 13, 7, 3 | 47, 48, 40, 30, 31, 32, 22, 14 | 30, 40, 48, 54, 58, 60 | 30, 29 |
| 50, 42, 32, 22, 14, 8, 7, 3 | 47, 39, 29, 19, 11, 12, 13, 14 | 30, 40, 48, 54, 58, 60 | 30, 29 |
| 50, 42, 32, 22, 14, 8, 7, 3 | 47, 39, 29, 30, 31, 21, 13, 14 | 30, 40, 48, 54, 58, 60 | 30, 29 |
| 50, 42, 32, 22, 14, 8, 7, 3 | 47, 39, 29, 30, 31, 21, 13, 14 | 30, 40, 48, 54, 58, 60 | 30, 29 |
| 50, 42, 32, 22, 14, 13, 7, 3 | 47, 39, 29, 19, 11, 12, 13, 14 | 30, 40, 48, 54, 58, 60 | 30, 29 |
| 50, 42, 32, 31, 21, 13, 7, 3 | 47, 48, 40, 30, 31, 32, 22, 14 | 30, 40, 48, 54, 58, 60 | 30, 29 |
| 50, 42, 32, 22, 14, 13, 7, 3 | 47, 39, 29, 19, 11, 12, 13, 14 | 30, 40, 48, 54, 58, 60 | 30, 29 |
| 50, 42, 32, 22, 14, 8, 7, 3 | 47, 39, 29, 30, 31, 21, 13, 14 | 30, 40, 48, 54, 58, 60 | 30, 29 |
| 50, 42, 32, 22, 14, 13, 7, 3 | 47, 39, 29, 19, 11, 12, 13, 14 | 30, 40, 48, 54, 58, 60 | 30, 29 |

(b)

**Fig. A2**. Generated paths for the Algorithm 1in Environment 2 for (a) the 2-sections robot, (b) the 4-sections robot.
Each line shows the generated path for a particular criteria combination group ("Combination" from the Table 3). Same paths are highlighted with the same colours.
The comparison ng both columns clearly shows that GA provides a higher variety of paths than A*.